\documentclass[10pt,twocolumn,letterpaper]{article}

\usepackage[pagenumbers]{cvpr} 

\definecolor{cvprblue}{rgb}{0.21,0.49,0.74}
\usepackage[pagebackref,breaklinks,colorlinks,allcolors=cvprblue]{hyperref}

\def\paperID{5497} 
\def\confName{CVPR}
\def\confYear{2025}

\begin{document}

\title{Dense Dispersed Structured Light \\for Hyperspectral 3D Imaging of Dynamic Scenes}
\author{
Suhyun Shin \\
POSTECH\\
\and
Seungwoo Yoon \\
POSTECH\\
\and
Ryota Maeda \\
POSTECH, University of Hyogo\\
\and
Seung-Hwan Baek \\
POSTECH 
}

\twocolumn[{
\renewcommand\twocolumn[1][]{#1}
\maketitle
\vspace{-20pt}
\begin{center}
    \centering
    \captionsetup{type=figure}
    \includegraphics[width=\linewidth]{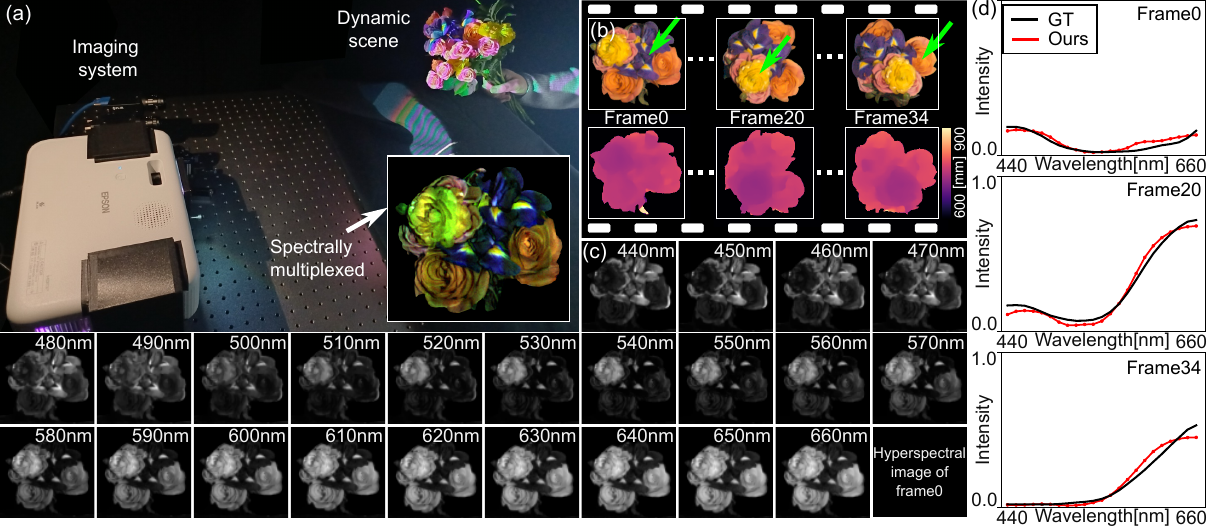}
    \vspace{-10pt}
    \captionof{figure}{We introduce a spectrally multiplexed Dense Dispersed Structured Light (DDSL), accurate hyperspectral 3D imaging method for dynamic scenes. (a) Capture configuration, (b) estimated hyperspectral image in sRGB and depth image for dynamic scenes, (c) estimated hyperspectral image, (d) comparison with spectroradiometer measurements.
    }
    \label{fig:teaser}
\end{center}
}]

\begin{abstract}
Hyperspectral 3D imaging captures both depth maps and hyperspectral images, enabling comprehensive geometric and material analysis. Recent methods achieve high spectral and depth accuracy; however, they require long acquisition times—often over several minutes—or rely on large, expensive systems, restricting their use to static scenes. We present Dense Dispersed Structured Light (DDSL), an accurate hyperspectral 3D imaging method for dynamic scenes that utilizes stereo RGB cameras and an RGB projector equipped with an affordable diffraction grating film.
We design spectrally multiplexed DDSL patterns that significantly reduce the number of required projector patterns, thereby accelerating acquisition speed. Additionally, we formulate an image formation model and a reconstruction method to estimate a hyperspectral image and depth map from captured stereo images. As the first practical and accurate hyperspectral 3D imaging method for dynamic scenes, we experimentally demonstrate that DDSL achieves a spectral resolution of 15.5 nm full width at half maximum (FWHM), a depth error of 4 mm, and a frame rate of 6.6 fps.
\end{abstract}

\begin{figure*}[t]
	\centering
		\includegraphics[width=\linewidth]{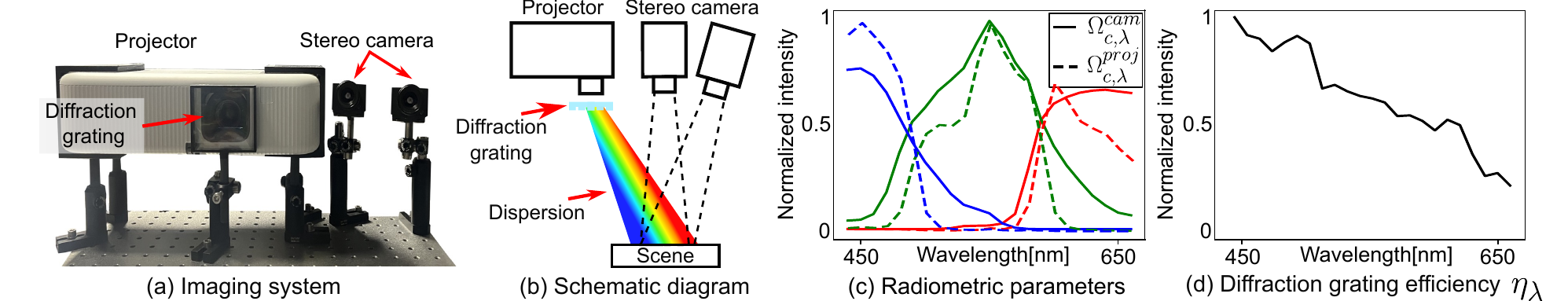}
  \caption{\textbf{Imaging System.} (a) Our active stereo system comprises RGB stereo cameras and a RGB projector equipped with a diffraction grating. (b) The diffraction grating introduces dispersion to the projector light. (c) Spectral sensitivity and emission functions of the camera and the projector. (d) {Diffraction grating efficiency.} }
  \vspace{-1mm}
		\label{fig:imaging_sys}
\end{figure*}

\section{Introduction}
\label{sec:intro}
Hyperspectral imaging captures a scene across multiple spectral channels beyond the three-channel RGB imaging, thereby providing both spectral and spatial information about the scene. Expanding on this, hyperspectral 3D imaging has recently received interest, as it captures both spectral and geometric information in the form of a depth map and a hyperspectral image. It has proven useful in scientific analysis of real-world materials, with applications in object classification~\cite{mahesh2015hyperspectral}, food ripeness detection~\cite{sun2010hyperspectral}, cultural heritage analysis, and geology~\cite{kim20123d}. Recently, dispersed structured light has been proposed as a cost-effective and compact solution for accurate hyperspectral 3D imaging~\cite{shin2024dispersed}. However, its applicability remains limited to static scenes due to the need for projecting hundreds of patterns, resulting in acquisition times of several minutes.

Increasing the acquisition speed of hyperspectral 3D imaging could make it feasible to analyze the geometric and material properties of objects and scenes in motion. Existing methods rely on expensive, bulky systems, such as coded-aperture snapshot spectral imagers {(CASSI)~\cite{wagadarikar2008single} paired with stereo cameras or time-of-flight setups~\cite{wang2015high, rueda2019snapshot, wang2016simultaneous, heist20185d, xiong2017snapshot}.} While compact, practical PSF-based systems exist~\cite{shi2024split, baek2021single}, they significantly compromise either depth or spectral accuracy.

In this paper, we propose DDSL, an accurate hyperspectral 3D imaging method for dynamic scenes, using a compact and affordable system. Figure~\ref{fig:teaser} shows the capture configuration using our prototype, which consists of stereo RGB cameras and an RGB projector equipped with an affordable diffraction-grating film that generates structured-light projections with wavelength-dependent dispersion.

We design DDSL patterns for the projector that produce spectrally multiplexed light projections, allowing us to use fewer than ten projections, enabling rapid image acquisition for dynamic scenes. We analyze the stereo images captured under these repeating DDSL patterns by developing an image formation model and a reconstruction method for depth maps and hyperspectral images.

DDSL enables accurate depth and spectral estimation even for high-frequency spectral variations, where existing affordable methods fall short. We demonstrate that DDSL achieves a depth error of 4 mm, a spectral FWHM of 15.5 nm in the visible spectrum, and acquisition speeds of 6.6 FPS. The use of a compact and affordable active-stereo setup enhanced with a diffraction grating film makes DDSL a promising approach for practical and accurate hyperspectral 3D imaging of dynamic scenes.

We summarize our contributions as follows:
\begin{itemize}
\item We introduce Dense Dispersed Structured Light (DDSL), which enables high-quality hyperspectral 3D imaging for dynamic scenes using an affordable active-stereo setup composed of RGB stereo cameras and an RGB projector augmented with a diffraction grating film.
\item We design DDSL patterns and develop an image formation model and hyperspectral 3D reconstruction method for dynamic scenes, obtaining a depth map and a hyperspectral images from stereo RGB images.
\item We demonstrate that DDSL outperforms state-of-the-art affordable hyperspectral 3D imaging methods in acquisition speed with high reconstruction accuracy, achieving a depth error of 4 mm, a spectral FWHM of 15.5 nm, and a frame rate of 6.6 FPS.
\end{itemize}
\section{Related Work}
\label{sec:related}

\paragraph{Hyperspectral 3D Imaging}
Various hyperspectral 3D imaging systems have been developed. Kim et al.\cite{kim20123d} combined a laser 3D scanner with a CASSI system for high-accuracy hyperspectral 3D imaging, while Li et al.\cite{li2019pro} employed a practical projector-camera setup, though at the cost of reduced spectral accuracy. Shin et al.~\cite{shin2024dispersed} introduced a compact projector-camera system using dispersed structured light with a diffraction grating. However, these previous approaches are generally limited to static scenes due to their long acquisition times, often lasting several minutes.
For dynamic scenes, a common approach is to use depth video cameras in conjunction with hyperspectral video cameras. However, hyperspectral video cameras tend to be large and expensive~\cite{cao2011prism, wang2015high, rueda2019snapshot, wang2016simultaneous, arablouei2016fast, bachmann2018low, van2010tracking, heist20185d}. While solutions using point spread function (PSF) engineering offer single-shot hyperspectral 3D imaging in a compact setup through custom micro- or nano-optical elements, they generally have limited spectral and depth accuracy~\cite{baek2021single, shi2024split}. Our method achieves accurate hyperspectral 3D imaging for dynamic scenes using a practical setup with stereo RGB cameras and an RGB projector equipped with a diffraction grating film.

\paragraph{Active Stereo}
Active stereo systems employ a stereo camera and an illumination module projecting structured-light patterns for robust 3D imaging~\cite{bleyer2011patchmatch, hirschmuller2007stereo, zhang2018activestereonet, riegler2019connecting, baek2021polka}. Although traditionally used for 3D imaging, Heist et al.~\cite{heist20185d} developed an active stereo system with a high-speed projector and hyperspectral video cameras for hyperspectral 3D imaging of dynamic scenes. However, the use of hyperspectral video cameras significantly increases instrumentation costs. Our approach leverages a diffraction grating film in front of an RGB projector with RGB stereo cameras, eliminating the need for hyperspectral video cameras and enabling practical, accurate hyperspectral 3D imaging for dynamic scenes.

\paragraph{Dispersive Optics}
Dispersive optics, including prisms and diffraction gratings, are widely used in hyperspectral imaging. CASSI systems utilize relay lenses with dispersive elements and coded masks for precise spectral reconstruction. Cao et al.~\cite{cao2011prism} minimized system size by using a prism and coded mask without relay lenses. Recently, diffractive optical elements have been employed to create spectrally varying point spread functions~\cite{baek2021single, shi2024split}. Shin et al.~\cite{shin2024dispersed} integrated a diffraction grating film into a projector-camera system, though it required hundreds of projected patterns, resulting in long capture times. Our method achieves rapid acquisition at 6.6 fps with high spectral and depth accuracy, through our active stereo setup, DDSL patterns, image formation model, and reconstruction method.
\section{Imaging System}
\label{sec:system}
We introduce a practical and affordable active stereo system. 
We use stereo RGB cameras (FLIR GS3-U3-32S4C-C) and an RGB projector (Epson CO-FH02) equipped with a thin diffraction grating film (Edmund 54-509) placed in front of the projector (see Figure~\ref{fig:imaging_sys}(a)). The diffraction grating film, which costs less than 20\,USD,  disperses the broadband projector light according to the light wavelength $\lambda$ as shown Figure~\ref{fig:imaging_sys}(b).
Each dispersed light ray of a specific wavelength $\lambda$ then propagates to a scene, creating spatially-distributed narrow-band spectral illumination. 
The stereo cameras capture the scene illuminated by the dispersed light. 
We set the camera fields of view to capture first-order diffracted light~\cite{Hecht:2001:Optics}.
For color channel $c \in \{R,G,B\}$ and wavelength $\lambda$, {we calibrate and refine} the projector spectral emission $\Omega^{\text{proj}}_{c, \lambda}$, camera spectral sensitivity $\Omega^{\text{cam}}_{c, \lambda}$, and diffraction-grating efficiency $\eta_{\lambda}$ as shown in Figures~\ref{fig:imaging_sys}(c) and (d).

\section{Image Formation}
\label{sec:image_formation}
We develop an image formation model for the active stereo system given a pattern image $P$ that we set to the projector. 

\paragraph{Projector Light}  
Given the projector pattern $P(q, c)$ where $q$ is a projector pixel and $c \in \{R, G, B\}$ is a color channel, we model the light intensity $L(q, \lambda)$ emitted from the pixel $q$ at wavelength $\lambda$ as
\begin{equation}
\label{eq:Light_intensity_L}
L(q, \lambda) = \sum_{c} \Omega^{\text{proj}}_{c, \lambda} \, P(q, c).
\end{equation}

\begin{figure}[t]
    \centering
    \includegraphics[width=1\linewidth]{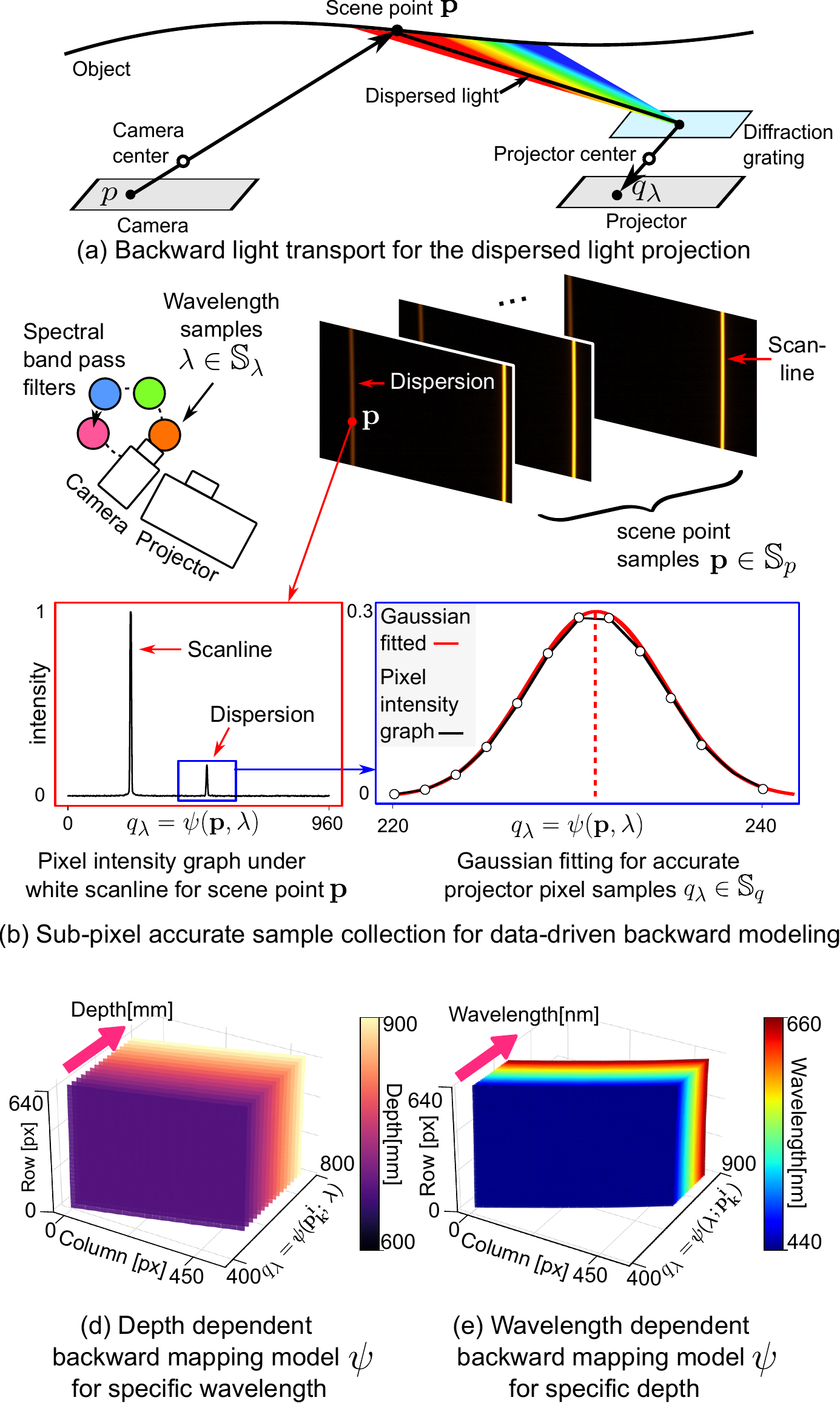}
    \caption{\textbf{Image Formation.} (a) {Light transport of the dispersed light projection of the mapping function $\psi$.} (b) Sub-pixel accurate sample collection for data-driven backward modeling. Calibrated backward mapping model that relates pixel point to projector horizontal position {(c) for depth given a specific wavelength and (d) for wavelength given a fixed depth value.}
    } 
    \label{fig:image_formation}
    \vspace{-2mm}
\end{figure}

\begin{figure*}[t]
	\centering
		\includegraphics[width=\linewidth]{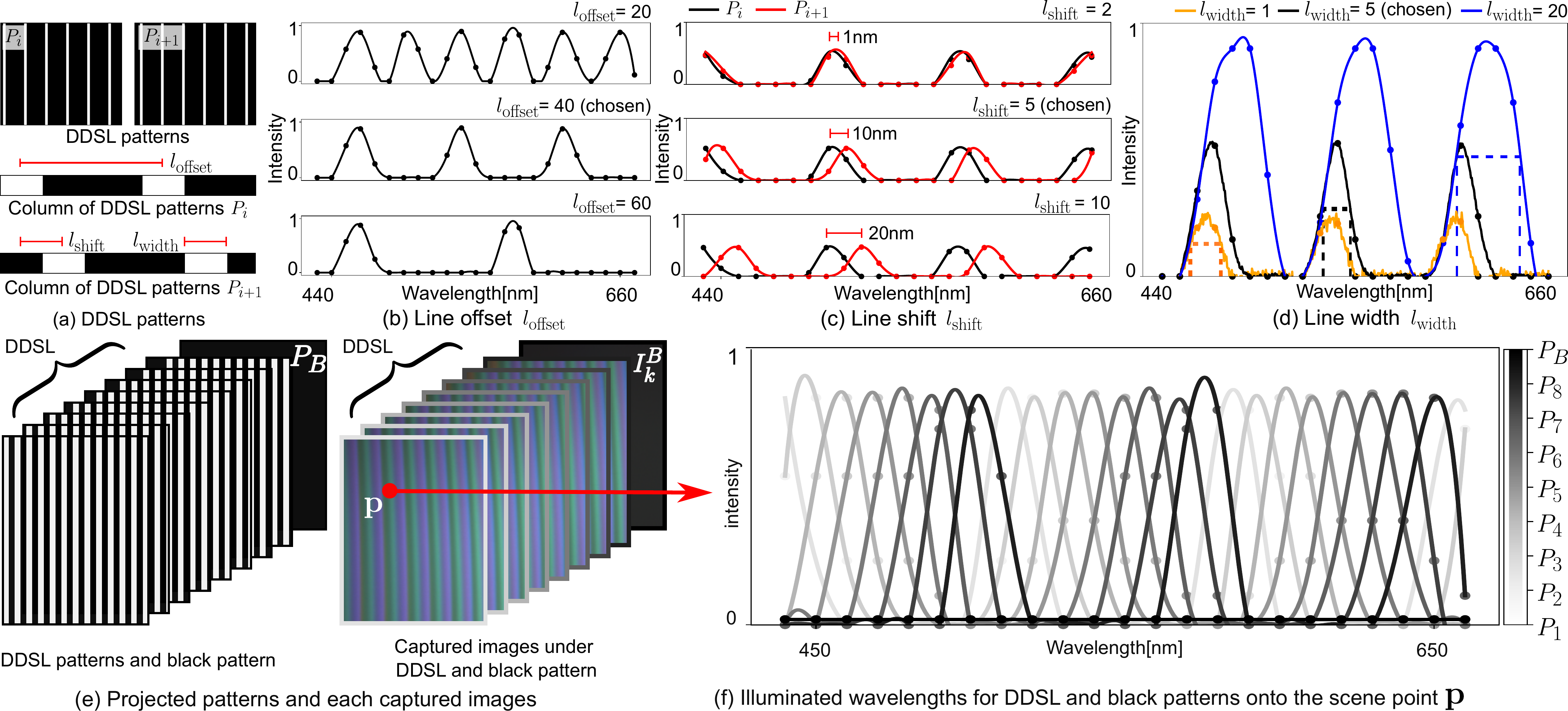}
  \caption{\textbf{DDSL Pattern Designs.} (a) Three parameters of DDSL patterns. We visualize the issues raised from large or small values for each parameter (b) line offset, (c) line shift, (d) line width. Refer to the text for details. (e) We project DDSL patterns and a single black pattern, and the captured images under such patterns are spectrally multiplexed. (f) Illuminated wavelengths for DDSL and black patterns onto the scene point.}
  \vspace{-2mm}
		\label{fig:design}
\end{figure*}

\paragraph{Dispersed Light Projection Model}
The light ray of wavelength $\lambda$ emitted by the projector pixel $q$ is diffracted by the diffraction grating and propagates to a scene. We model such dispersed projection in a backward manner as shown Figure~\ref{fig:image_formation}(a). That is, from a scene point $\mathbf{p}$ and given wavelength $\lambda$, we model its corresponding projector pixel $q_{\lambda}$ that emits the ray:
\begin{equation}
    \label{eq:correspondence}
    q_{\lambda} = \psi\left(\mathbf{p},\lambda\right).
\end{equation}
To construct the backward model $\psi$, we use a data-driven approach. First, we acquire the samples ${q_{\lambda} \in \mathbb{S}_q, \lambda \in \mathbb{S}_{\lambda}, \mathbf{p} \in \mathbb{S}_{\mathbf{p}}}$ by capturing Spectralon images under column-wise scan-line patterns per each narrow-band spectrum $\lambda$ using spectral bandpass filters as shown in Figure~\ref{fig:image_formation}(b). Using the scan-line patterns and sub-pixel accurate Gaussian fitting, we obtain the scene point $\mathbf{p}$ via triangulation, and its projector pixel $q_{\lambda}$ can be obtained for the horizontal coordinate. 
Note that we model the horizontal coordinate only as dispersion occurs in the horizontal direction.
Second, we apply non-linear interpolation using a power function along the depth coordinate of the scene point $\mathbf{p'}_{z} \in \mathbb{S}_{\mathbf{p}}$, where $\mathbf{p'}_{z}$ denotes the z-coordinate of a scene point $\mathbf{p'}$, and then linearly interpolate wavelength samples $\lambda' \in \mathbb{S}_{\lambda}$ and spatial samples $p' \in \mathbb{S}_p$. As a result, we obtain the mapping function $\psi$ with a sub-pixel reprojection error of 0.66\, pixel. 
More details of the sample acquisition and interpolation can be found in the Supplemental Document.

\begin{figure*}[t]
	\centering
		\includegraphics[width=\linewidth]{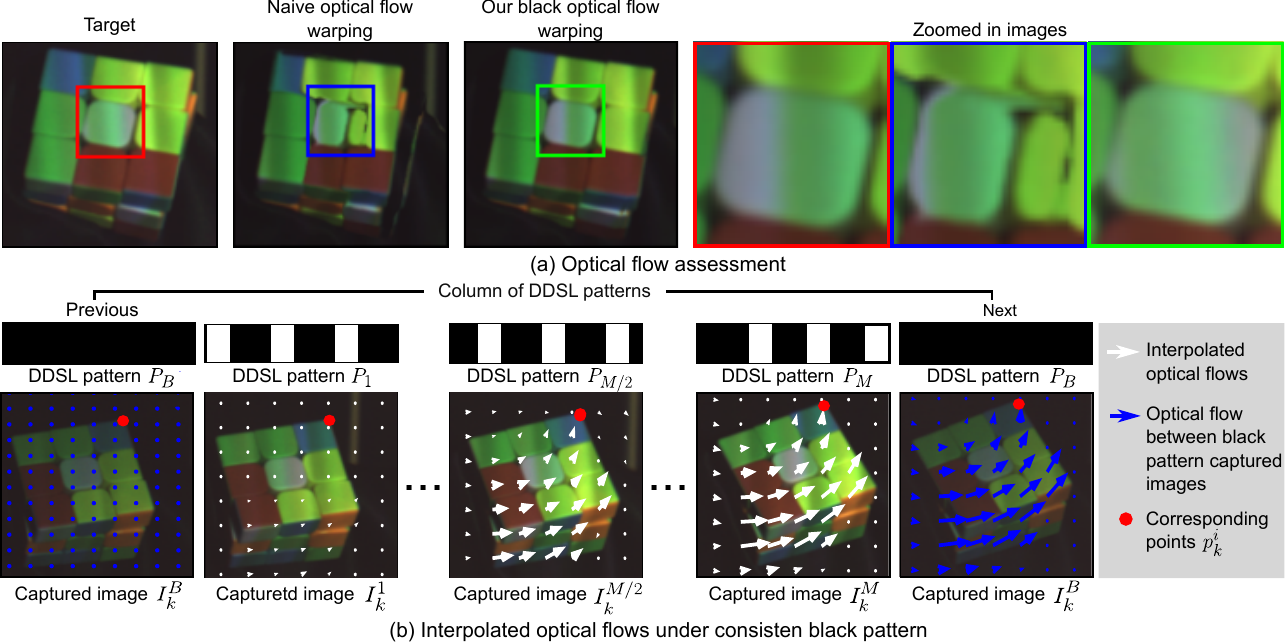}
  \caption{\textbf{Black Optical Flow.}
  We warp adjacent frame image to a target frame using naive optical flow under DDSL patterns and our black optical flow method to show the effectiveness of our method. We estimate optical flows using pretrained RAFT~\cite{lipson2021raft} network for both methods. Note that the target frame and adjacent frames are captured under different DDSL patterns. Therefore, the evaluation should primarily focus on geometric alignment rather than color consistency. (a) Warped images based on naive optical flow and black optical flow. (b) Visualization of the interpolated optical flows using the black optical flow.}
  
  \vspace{-2mm}
		\label{fig:optical_flow}
\end{figure*}

\paragraph{Stereo Imaging} 
The camera $k \in \{\text{left}, \text{right}\}$ captures the scene point $\mathbf{p}$ at a camera pixel $p_k$ through perspective projection:
\begin{equation}
\mathbf{p} = z_k\mathbf{E}_k^{-1}\mathbf{K}_k^{-1}\dot{p}_k,
\end{equation}
where $z_k$ is the depth, $\mathbf{E}_k$ is the extrinsic matrix, $\mathbf{K}_k$ is the intrinsic matrix, and $\dot{p}_k$ is the homogeneous coordinate of pixel $p_k$.
The captured intensity $I_k(p_k, c)$ for the color channel $c$ is modeled as
\begin{align}
\label{eq:image_formation}
I_k(p_k,c) = \sum_{\lambda \in \Lambda} \Omega^{\text{cam}}_{c, \lambda}  H_k(p_k,\lambda) \frac{\eta_{\lambda}}{d(\mathbf{p})^2}  L(q_{\lambda}, \lambda),
\end{align}
where $H_k(p_k, \lambda)$ is the hyperspectral image and $d(\mathbf{p})$ is the distance between the scene point $\mathbf{p}$ and the projector. The model integrates over wavelengths $\lambda \in \Lambda$, effectively ranging from 440\,nm to 660\,nm at 10\,nm intervals: $\Lambda=\{\lambda_{1}=440\,nm, \cdots, \lambda_{N}=660\,nm\}$, where $N=23$ is the number of spectral bands.

\section{Dense Dispersed Structured Light}
\label{sec:dense_dispersed_sl}
We design DDSL patterns $\{P_i\}_{i=1}^{M}$ that enable hyperspectral 3D imaging for dynamic scenes on our active stereo setup. 
Each DDSL pattern $P_i$ is composed of multiple vertical lines, defined as:
\begin{equation}
    P_{i}(q, \forall) = 
    \begin{cases}
        1 & \text{if} \quad  \operatorname{mod}(|q_x - i \times l_{\text{shift}}|, l_{\text{offset}})  \leq \frac{l_{\text{width}}}{2} , \\
        0 & \text{else,}
    \end{cases}
    \label{eq:projector_pattern_P}
\end{equation}
where $q_x$ is the column index of a projector pixel $q$, $l_\text{shift}$ is the line shift between neighboring patterns $P_i$ and $P_{i+1}$, $l_\text{offset}$ is the line offset between neighboring lines in a pattern, $l_\text{width}$ is the line width, and $\operatorname{mod}(x,y) = x \% y$ is the modulo operator.

We project $M$ DDSL patterns $\{P_1, \dots, P_M\}$ and  one black pattern $P_B$ repeatedly, capturing corresponding stereo images $I^1_k, \dots, I^M_k, I^B_k$ for $k \in \{\text{left}, \text{right}\}$. 
The black pattern $P_B$ is used to compensate for the non-zero projector intensity for the zero-valued pattern and to enable robust motion compensation which we detail in Section~\ref{sec:image_formation_multi}.
The DDSL patterns create dispersion for each vertical white line, and the dispersed patterns can overlap depending on the settings for three parameters: line offset $l_\text{offset}$, line width $l_\text{width}$, and line shift $l_\text{shift}$.
Figure~\ref{fig:design}(a) and (e) shows the DDSL patterns and captured images with our chosen parameters  $l_\text{offset}$, $l_\text{width}$, and $l_\text{shift}$.
 Below, we discuss our design choices for these parameters.

\paragraph{Line Offset}
The line offset $l_\text{offset}$ defines the spacing between vertical lines in pattern $P_i$. A small $l_\text{offset}$ increases overlap between dispersed patterns, resulting in more spectrally multiplexed illumination per scene point (Figure~\ref{fig:design}(b)). This reduces the number of required patterns $M$, though it may decrease spectral accuracy due to blurred illumination. A larger $l_\text{offset}$ reduces multiplexing, improving spectral accuracy at the cost of a higher number of patterns. We set $l_\text{offset} = 40\,\text{px}$ to multiplex three spectral bands aligned with the camera RGB channels, balancing spectral accuracy and pattern count.

\begin{figure*}[t]
	\centering
		\includegraphics[width=\linewidth]{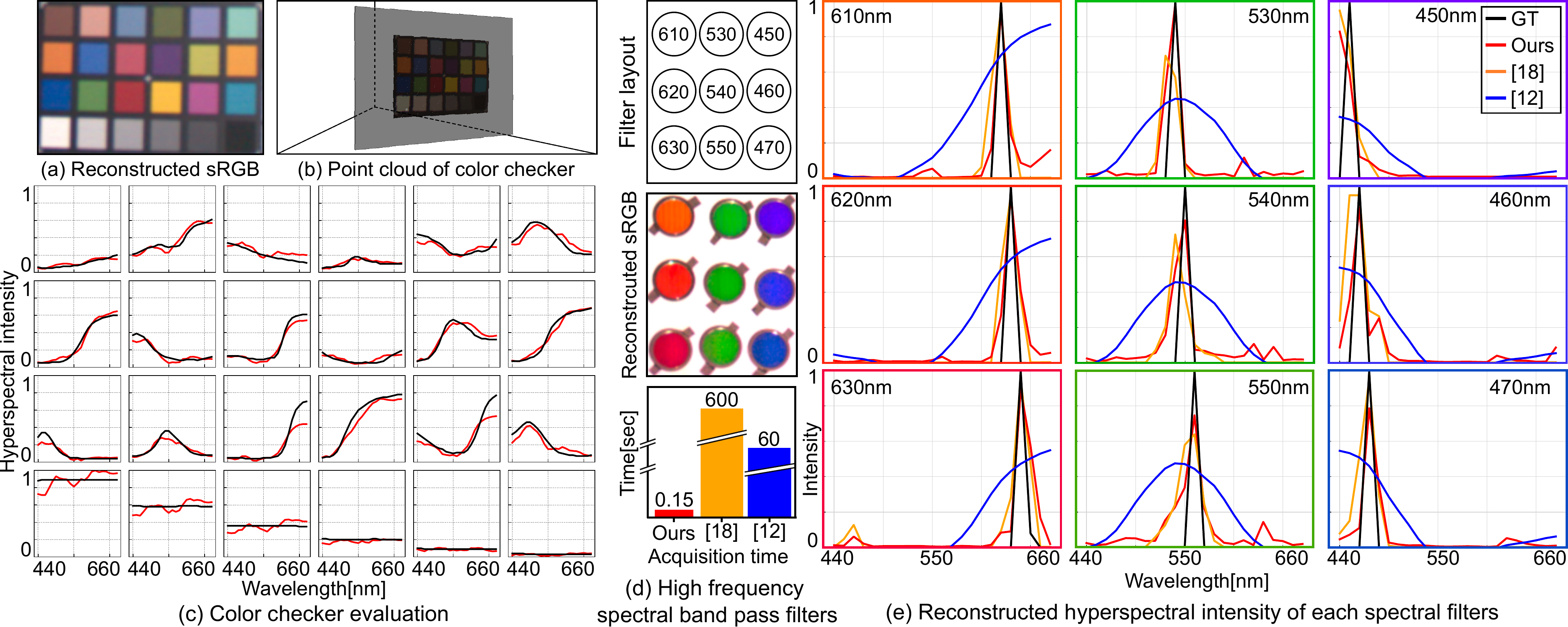}
  \caption{\textbf{Spectral Accuracy.} Color checker evaluation. (a) Reconstructed hyperspecral image of color checker in sRGB, (b) point cloud of color checker, (c) hyperspectral intensity of color checker. {High-frequency spectral data evaluation comparing other methods with Shin et al.~\cite{shin2024dispersed} and Li et al.~\cite{li2019pro}. (d) 10\,nm-FWHM spectral band pass filters and acquisition time for the three methods, (e) reconstructed hyperspectral intensity.}}
  \vspace{-2mm}
		\label{fig:high_freq}
\end{figure*}

\paragraph{Line Shift}
The line shift $l_\text{shift}$ specifies the shift between lines in neighboring patterns $P_i$ and $P_{i+1}$. A small $l_\text{shift}$ densely samples the spectral axis, increasing potential spectral channels but requiring more patterns $M$ (Figure~\ref{fig:design}(c)). To achieve spectral channels from 440\,nm to 660\,nm at 10\,nm intervals, we set $l_\text{shift} = 5\,\text{px}$, providing a 10\,nm spectral step size while minimizing the number of patterns.

\paragraph{Line Width}
A larger line width $l_\text{width}$ increases spectral overlapping across RGB channels, smoothing the illumination spectrum and potentially lowering spectral reconstruction accuracy (Figure~\ref{fig:design}(d)). A very narrow $l_\text{width}$ results in low illumination power, introducing noise in captured images. We set $l_\text{width} = 5$ pixels to optimize illumination intensity without sacrificing spectral accuracy.

\paragraph{Summary}
In summary, we use {eight} DDSL patterns ($M=8$) with $l_\text{shift} = 5$, $l_\text{width} = 5$, and $l_\text{offset} = 40$, enabling accurate hyperspectral 3D reconstruction in our experimental setup.
Figure~\ref{fig:design}(f) shows the spectral power distribution of the DDSL-pattern illumination {projected onto} a scene point $\mathbf{p}$. 
For each $i$-th DDSL pattern, we have three peaks over the RGB spectrum. Using only  $M=8$ DDSL patterns, we can densely sample wavelengths with 10\,nm step, allowing for accurate hyperspectral 3D imaging, contrast to using hundreds of pattern images as in the previous work~\cite{shin2024dispersed}.
\section{Hyperspectral 3D Reconstruction}
\label{sec:reconstruction}
Given the captured stereo images $I^1_k, \dots, I^M_k, I^B_k$ for DDSL patterns $P_1, \dots, P_M$ and one black pattern $P_B$, we reconstruct a hyperspectral image $H_k$ and a depth map $z_k$. 

\paragraph{Depth Estimation} 
We estimate depth from stereo images $I^i_\text{left}$ and $I^i_\text{right}$ for each frame $i$. The captured stereo images are rectified, and disparity is estimated using the pretrained RAFT-Stereo network~\cite{lipson2021raft}. Depth is then computed from disparity using calibrated camera parameters, followed by un-rectification. 

\paragraph{Black Optical Flow}
\label{sec:image_formation_multi}
Dynamic objects move while capturing the images $I^1_k, \dots, I^M_k, I^B_k$ under the varying projector patterns $P_1, \dots, P_M, P_B$, which needs compensation for robust hyperspectral image reconstruction. One straightforward approach is to estimate optical flow $\nabla p_k^{M/2 \to i}$ from the center frame $M/2$ to each frame $i$:
\begin{equation}
    p_k^i = p_k^{M/2} + \nabla p_k^{M/2 \to i}.
\end{equation}
However, this approach is challenging due to the inconsistent illumination at corresponding pixels since each image $\{I^i_k\}_{i=1}^M$ is illuminated by a different DDSL pattern $\{P_i\}_{i=1}^M$. {Figure~\ref{fig:optical_flow}(a) depicts how naive optical flow between adjacent DDSL patterns fails to accurately capture the motion.} Instead, we estimate the black optical flow, which is the flow between successive black-pattern images $I^B_k$, avoiding illumination inconsistency. We then interpolate the black optical flow to obtain the target optical flow $\nabla p_k^{M/2 \to i}$, enabling robust optical-flow estimation, as illustrated in Figure~\ref{fig:optical_flow}.
Using the estimated flow, we align the images $I^1_k, \dots, I^M_k$ captured under the DDSL patterns to the center frame $I^{M/2}_k$.

\paragraph{Multi-pattern Image Formation}
Using the aligned images $I^1_k, \dots, I^M_k$ captured under $M$ DDSL patterns, we reformulate the image formation model from Section~\ref{sec:image_formation}. 
For the $i$-th DDSL pattern, the aligned image $I_k^i$ is modeled as:
\begin{align}
\label{eq:image_formation_dynamic}
    I_k^i(p_k^i, c) = \sum_{\lambda \in \Lambda} \Omega^{\text{cam}}_{c, \lambda} \, H_k(p_k^i, \lambda) \, \frac{\eta_{\lambda}}{d(\mathbf{p}_k^i)^2} \, L^i(q_\lambda^i, \lambda),
\end{align}
where $\mathbf{p}_k^i$ is the corresponding scene point at frame $i$, and $q_\lambda^i$ is the corresponding projector pixel obtained using the backward model $\psi$: $q_\lambda^i = \psi\left( \mathbf{p}_k^i, \lambda \right)$.

We then subtract the black pattern captured image $I^B_k$ from all the captured images $\{I_k^i(p^i_k, c)\}_{i=1}^M$ to remove the undesired residual light intensity for the black pattern:
\begin{align}
\label{eq:black_sub}
    I_k^i(p^i_k, c) \gets I_k^i(p^i_k, c) - I_k^B(p_k^B, c)
\end{align}

We rewrite the image formation as a matrix-vector multiplication:
\begin{equation}
\label{eq:image_formation_matrix}
    \mathbf{I}_k = \mathbf{L} \mathbf{H}_k,
\end{equation}
where $\mathbf{I}_k$ is the intensity vector of captured image under $M$ DDSL patterns, $\mathbf{L}$ is the system matrix, and $\mathbf{H}_k$ is the hyperspectral image:
\begin{align}
    \label{eq:I}
    \mathbf{I}_k =& [I_k^1(p_k^1, R), \dots, I_k^M(p_k^M, R), \nonumber \\
                  & \ I_k^1(p_k^1, G), \dots, I_k^M(p_k^M, G), \nonumber \\
                  & \ I_k^1(p_k^1, B), \dots, I_k^M(p_k^M, B)]^\intercal \in \mathbb{R}^{3M \times 1},\\
    \mathbf{L} =& \begin{bmatrix} \mathbf{L}_R; \mathbf{L}_G; \mathbf{L}_B \end{bmatrix}^\intercal \in \mathbb{R}^{3M \times N}, \\ 
    \mathbf{L}_c(i, j) =& \Omega^{\text{cam}}_{c, \lambda_j} \, \eta_{\lambda_j} \, \left( L^i * G \right)\left( q_{\lambda_j}^i, \lambda_j \right),\\
    \mathbf{H}_k =& [H(p_k^{M/2}, \lambda_1), \dots, H(p_k^{M/2}, \lambda_N)]^\intercal \in \mathbb{R}^{N \times 1},
\end{align}

where $i = 1, \dots, M$, $j = 1, \dots, N$, and $G$ is a Gaussian blur kernel accounting for the imaging system's blur; details can be found in the Supplemental Document. 

\begin{figure}[t]
    \centering
    \includegraphics[width=\linewidth]{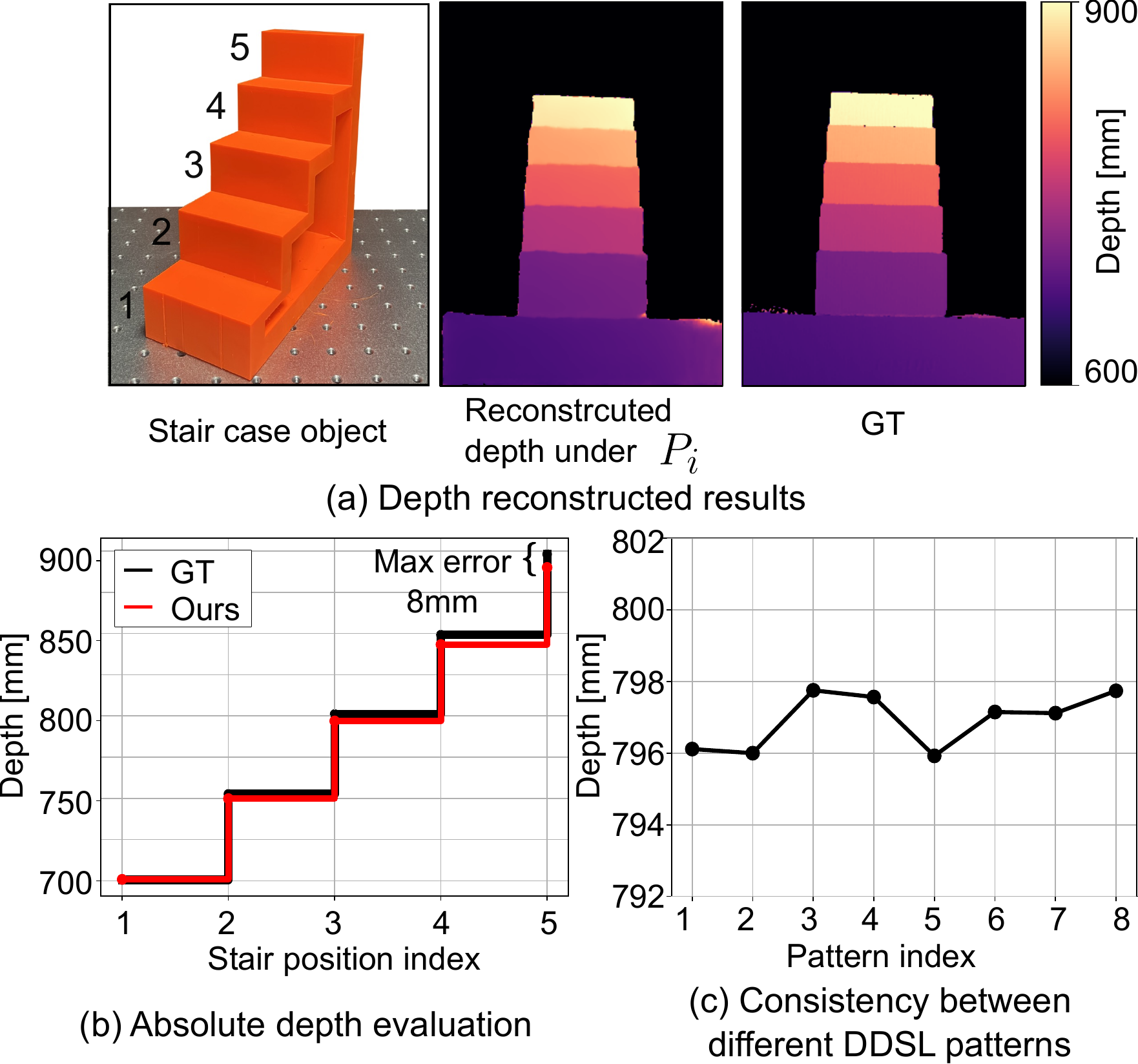}
    \caption{\textbf{Depth Validation.} (a) Stair case depth reconstructed results. (b) Absolute depth evaluation. (c) Consistency between depth reconstructed results between different DDSL patterns {for the stair position index of 3.}}
    \label{fig:depth_validation}
    \vspace{-3mm}
\end{figure}

\paragraph{Hyperspectral Reconstruction}
We perform per-pixel hyperspectral reconstruction $\mathbf{H}_k$ by solving the following optimization problem:
\begin{align}
    \label{eq:optimization}
     \underset{\mathbf{H}_k}{\text{argmin}} \,
    &\underbrace{\| \mathbf{L} \mathbf{H}_k - \mathbf{I}_{k} \|_2^2}_{\text{Data term}}  +
    \underbrace{\kappa_\lambda \| \nabla_\lambda \mathbf{H}_k \|_2^2}_{\text{Spectral smoothness}} + \nonumber \\
    &\underbrace{\kappa_{xy} \left( \| \nabla_x \mathbf{H}_k \|_1 + \| \nabla_y \mathbf{H}_k \|_1 \right)}_{\text{Spatial regularization}},
\end{align}
where $\nabla_\lambda$, $\nabla_{x}$, and $\nabla_{y}$ are gradient operators along the spectral and spatial axes, respectively. The first term corresponds to the data term, penalizing reconstruction error, while the second and third terms enforce spectral smoothness and spatial total variation. The coefficients $\kappa_{\lambda}=3$ and $\kappa_{xy}=0.05$ are balancing weights. We solve the per-pixel optimization problem using gradient descent in PyTorch. Details of the optimization are provided in the Supplemental Document.

\section{Calibration}
\label{sec:calibration}
We perform a one-time calibration of the projector, camera, and diffraction grating. We obtain the geometric parameters of the stereo cameras and the projector using checkerboard methods~\cite{taubin20143d,zhang2000flexible} without attaching the diffraction grating film. To measure the diffraction efficiency $\eta_\lambda$ of the diffraction grating, shown in Figure~\ref{fig:imaging_sys}(d), we filter the dispersed light using spectral bandpass filters with a 10\,nm bandwidth, covering wavelengths from 440\,nm to 660\,nm, and capture the intensity reflected from a Spectralon target.
{We calibrate the camera response function $\Omega^{\text{cam}}_{c,\lambda}$ also using spectral bandpass filters under LED light}, and measure the projector spectral emission function $\Omega^{\text{proj}}_{c,\lambda}$ by projecting RGB dots onto a Spectralon target and capturing the reflected radiance with a spectroradiometer (JETI Specbos 1211).
To enhance reconstruction accuracy, we further optimize the radiometric parameters for both the projector and camera; detailed methods are provided in the Supplemental Document.

\begin{figure}[t]
    \centering
    \includegraphics[width=\linewidth]{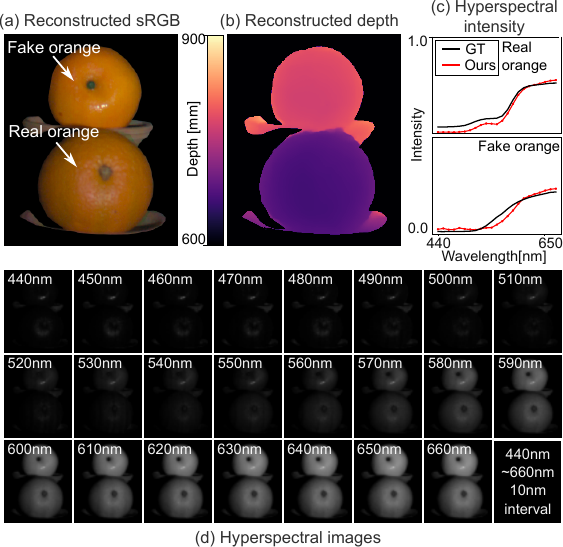}
    \caption{\textbf{Fake and Real Oranges}. We compare real and artificial fruit(orange), showing the differences in their hyperspectral image curves and corresponding images. (a) Reconstructed hyperspectral image in sRGB, (b) depth (c) spectral graph and (d) hyperspectral images of metameric samples.}
    \label{fig:metamerism}
    \vspace{-3mm}
\end{figure}
\begin{figure*}[t]
	\centering
		\includegraphics[width=\linewidth]{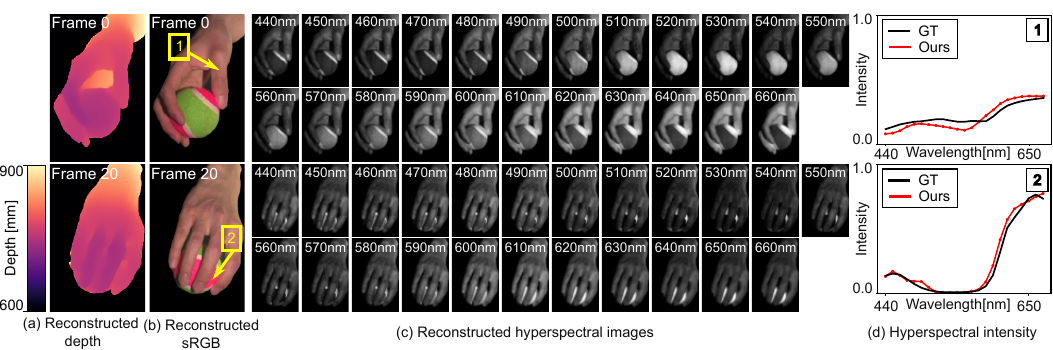}
  \caption{\textbf{Dynamic Scene.} We show the reconstructed depth and spectral results of a dynamic scene. (a) Reconstructed depth. (b) Reconstructed hyperspectral image in sRGB. (c) Reconstructed hyperspectral image. (d) Hyperspectral intensity.}
  \vspace{-2mm}
		\label{fig:hand}
\end{figure*}

\section{Results}
\label{sec:assessment}

\paragraph{Spectral Accuracy}
Figure~\ref{fig:high_freq} (a), (b) and (c) shows the reconstructed hyperspectral image in sRGB, depth, and hyperspectral curves of a ColorChecker chart, demonstrating accurate reconstruction. To measure spectral FWHM of our system, we capture nine narrow-band spectral filters as shown in Figure~\ref{fig:high_freq}(d).
The FWHM of those bandpass filters is 10\,nm.
Figure~\ref{fig:high_freq}(e) shows that our method achieves a FWHM of 15.5\,nm compared to the ground-truth hyperspectral data of the filters, outperforming state-of-the-art affordable hyperspectral 3D imaging methods by Li et al.~\cite{li2019pro} and Shin et al.~\cite{shin2024dispersed}.
Li et al.\cite{li2019pro} shows a FWHM of 40\,nm due to the use of broadband RGB illumination of a conventional projector. 
Shin et al.\cite{shin2024dispersed} attains 18\,nm FWHM. Moreover, its acquisition time for a scene is 10 minutes whereas our method only requires 0.15\,seconds, representing $\times$4000 speed increase, thus enabling hyperspectral 3D imaging for dynamic scenes.

\paragraph{Depth Accuracy} 
We evaluate the accuracy of depth estimation. Ground-truth depth is obtained using the binary-coded structured light method~\cite{geng2011structured}. 
Figure~\ref{fig:depth_validation}(a) evaluates the absolute depth error by capturing a 3D-printed stair object. We achieve an average depth error of 4\,mm in the area of each step compared to the ground truth, with a maximum error of 8\,mm. Figure~\ref{fig:depth_validation}(b) shows the difference between depth results for each DDSL pattern. The consistency across different DDSL patterns is evident, with a difference of less than 2\,mm between patterns. 

\paragraph{Dynamic Scenes} 
By using spectrally multiplexed DDSL patterns, we reduce the required number of projections from over hundred patterns with naive scanning~\cite{shin2024dispersed} to 8\, DDSL patterns. This enables accurate hyperspectral 3D imaging for dynamic scenes at 6.6\,fps, as shown in Figure~\ref{fig:teaser}, Figure~\ref{fig:hand}, and the Supplemental Video.

\paragraph{Real and Fake Oranges}
Figure~\ref{fig:metamerism} presents a comparison between artificial and real oranges with reconstructed hyperspectral images in sRGB and depth with spectral curves of each fruits. We can differentiate between the objects in detailed spectral analysis. Ground-truth intensity measurements are acquired using a spectroradiometer.

\section{Conclusion}
\label{sec:conclusion}

We have introduced DDSL, an accurate and compact method for hyperspectral 3D imaging for dynamic scenes. We use a conventional RGB stereo camera-projector system paired with a sub-millimeter diffraction grating, implemented as a practical experimental prototype. We design the DDSL patterns, generating spectrally-multiplexed illumination, enabling rapid and high-quality hyperspectral 3D imaging. Our method incorporates the dispersion-aware image formation model using the sub-pixel accurate backward mapping. Experimental results demonstrate that we outperform prior affordable methods in accuracy and also acquisition speed  with a depth error of 4\, mm and a spectral accuracy of 15.5\, FWHM. We envision that our DDSL method opens up new applications of geometric and material analysis for dynamic objects. 

\paragraph{Limitations and Future work}
While our method enables rapid, accurate, and practical hyperspectral 3D imaging for dynamic scenes, it is currently constrained to low-speed motion, achieving a frame rate of 6.6\, fps due to the slow software-based synchronization between the camera and projector. To address this limitation, hardware synchronization and a high-speed, affordable projector-camera system could significantly increase frame rates~\cite{yu2020dual}. The limited diffraction efficiency also restricts the range of scene positions. Thus, using a high efficiency diffraction grating for the first-order light would enhance the effective range.

{\small
\bibliographystyle{ieeenat_fullname}
\bibliography{references}
}

\end{document}


\begin{CJK}{UTF8}{gbsn} 

\title{Dense Dispersed Structured Light \\for Hyperspectral 3D Imaging of Dynamic Scenes}
\author{
Suhyun Shin \\
POSTECH\\
\and
Seungwoo Yoon \\
POSTECH\\
\and
Ryota Maeda \\
POSTECH, University of Hyogo\\
\and
Seung-Hwan Baek \\
POSTECH 
}

\maketitle

In this supplemental document, we provide additional results and details in support of our findings in the main manuscript. 

\tableofcontents

\newpage

\section{Experimental Prototype}
We list all parts used to build the experimental prototype
system in Table~\ref{tab:part_list}.

\begin{table*}[h]
\centering
\label{tab:table1}
\begin{tabular}{clcc}
\hline
\textbf{Item \#} & \textbf{Part description} & \textbf{Quantity} & \textbf{Model name} \\
\hline
1 & RGB Camera & 2 & FLIR GS3-U3-32S4C-137
C \\
2 & RGB Projector & 1 & Epson CO-FH02 \\
3 & Objective lens & 2 & Edmund \#33-303\\
4 & Diffraction grating sheet & 1 & Edmund 158 \#54-509 \\
5 & Holder & 1 & 3D printed  \\
\hline
\end{tabular}
\caption{Part list of our imaging system.}
\label{tab:part_list}
\end{table*}

\subsection{Geometric Calibration}
The geometric calibration of the camera and projector is preformed without using diffraction grating~\cite{taubin20143d}. Similarly, the stereo-camera system undergoes geometric calibration based on the approach proposed in~\cite{zhang2000flexible}. Both calibration procedures achieve an average sub-pixel reprojection error of 0.1 pixels, ensuring precise geometric alignment with high accuracy. Using the calibrated geometric parameters of camera and projector, we establish the correspondence between a camera pixel $p$ and a projector pixel $q$ through the operations \texttt{unproject}($\cdot$) and \texttt{project}($\cdot$), as defined below: 

\begin{align}
    \label{eq:cam_proj_corres}
    q = \texttt{project}\left(\texttt{unproject}\left(p, \mathbf{p}_z\right)\right),
\end{align}
where $\mathbf{p}_z$ represents the depth value of scene point $\mathbf{p}$ corresponding to camera pixel $p$.

\subsection{Initial Radiometric Calibration of Spectral/Emission Function}

\subsubsection{Camera Response Function}
This section outlines the radiometric calibration process for determining the spectral response and emission functions.
To compute the initial spectral response function of the camera, we employ a Spectralon target with a uniform reflectance of 99\% across all visible wavelengths. The target is illuminated by an LED light source, whose spectral power distribution is precisely measured using a spectroradiometer (JETI Specbos 1211) to provide the ground truth reference for the calibration.

\begin{figure}[h]
    \centering
    \includegraphics[width=\linewidth]{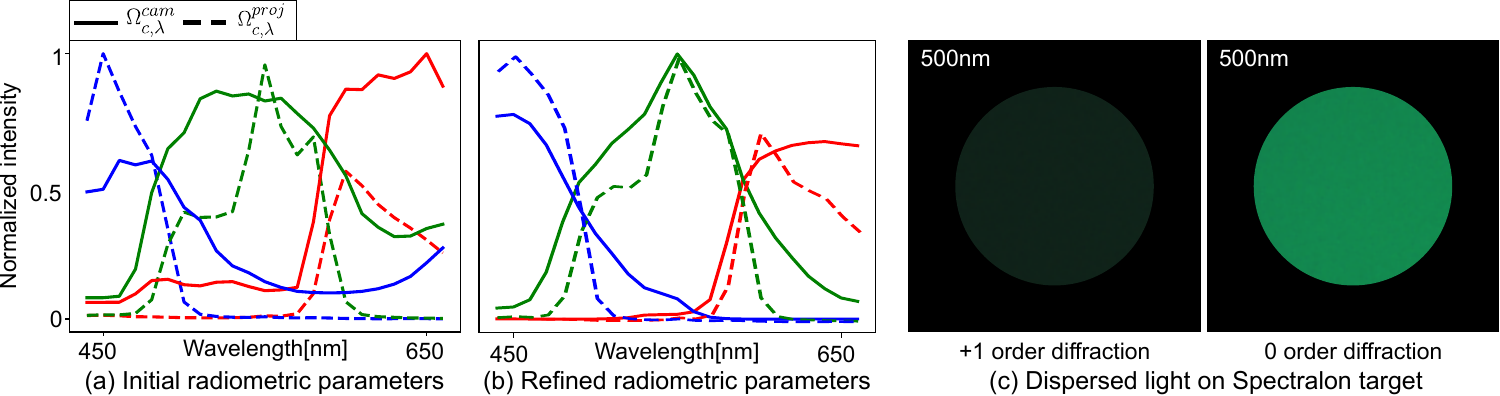}
    \caption{\textbf{Radiometric Calibration.} (a) Initial spectral and emission functions before optimization. (b) Refined radiometric parameters after optimization. (c) Captured dispersed light on the Spectralon target at a specific wavelength (500 nm) used for diffraction grating efficiency calibration.}
    \label{fig:radiometric_calibration}
\end{figure}

The camera's spectral response is defined over the wavelength range of 440 nm to 660 nm, sampled at 10 nm intervals. Narrow band spectral filters, corresponding to each interval, are sequentially placed in front of the camera. Each spectral images are captured under the same LED illumination for a total of 23 spectral bands. By analyzing the intensity values across these bandpass-filtered images, we compute the camera’s spectral response function, as expressed below:

\begin{align}
\label{eq:spectral_response_function}
    I_\lambda(p, c) = \sum_{\lambda \in \Lambda} \Omega^{\text{cam}}_{c, \lambda} \, H_{\text{Spectralon}}(p, \lambda) \, LED(\lambda),
\end{align}
where $I_{\lambda}(p, c)$ represents the intensity of the captured image at pixel $p$ within the region of interest (RoI) for spectral band $\lambda$ and RGB channel $c$. The term $H_{\text{Spectralon}}(p, \lambda)$ denotes the hyperspectral reflectance of the Spectralon target which is 99\%. The LED spectral power distribution, denoted as $LED(\lambda)$, serves as the ground truth for the light source, and $\Omega_{c, \lambda}^{\text{cam}}$ represents the spectral response function of the camera. We show the initial radiometric parameters in Figure~\ref{fig:radiometric_calibration}(a).

\subsubsection{Projector Emission Function}
Next, we calibrate the projector's emission function, denoted as $\Omega^{\text{proj}}_{c,\lambda}$ . This is achieved by projecting RGB patches onto the Spectralon target. The reflected spectral radiance of each projected patch is then measured using a spectroradiometer, allowing us to obtain the spectral emission curves for each of the RGB channels. The the initial radiometric parameter of projector is depicted in Figure~\ref{fig:radiometric_calibration}(a).

\subsection{Refinement of Spectral/Emission Function}
We refine the spectral response and emission functions using multiple white scanline patterns, similar to the approach proposed by Shin et al.~\cite{shin2024dispersed}. The refinement process minimizes the difference between the measured intensity $\mathbf{I}$, and the simulated intensity under scanline patterns, as expressed in Equation~\ref{eq:optimization_resp}. This refinement is based on the simulated pixel intensity graph for 21 centeral points $p \in \mathcal{P}$ of the Classic ColorChecker, using the known ground truth hyperspectral reflectance $H(p,\lambda)$.

\begin{equation}
        \underset{\mathbf{\Omega's}}{\text{argmin}} \sum_{p \in \mathcal{P}} \| \textbf{I}_k -
     \Omega^{\text{cam}}_{c, \lambda}  H_k(p,\lambda)\eta_{\lambda}L(q_{\lambda}, \lambda)\|_2^2
     + w(\| \nabla_\lambda \Omega^{\text{cam}}_{c, \lambda}\|_2^2 + \| \nabla_\lambda \Omega^{\text{proj}}_{c', \lambda}\|_2^2).
    \label{eq:optimization_resp}
\end{equation}

In this formulation, $\nabla_\lambda$ represents the gradient operator along the spectral axis, which ensures the smoothness of the spectral response functions. The regularization term, weighted by $w = 0.008$, penalizes variations in the spectral gradients of the both camera's spectral response function $\Omega^{\text{cam}}_{c,\lambda}$, and the projector's emission function $\Omega^{\text{proj}}_{c,\lambda}$. The results of the optimized spectral response and emission functions are presented in Figure~\ref{fig:radiometric_calibration}(b), and this refinement leads to a more accurate representation of the radiometric properties, enabling improved spectral and spatial reconstruction.

\subsection{Diffraction Grating Efficiency}
A diffraction grating disperses incident light into multiple diffraction orders. The zero-order diffraction preserves the original path of the incident light, maintaining the same intensity across all wavelengths. In contrast, the first-order diffraction, which is utilized in our system, separates the incident light into different angles based on wavelength. Unlike the zero-order light, the first-order diffraction exhibits varying efficiency across different wavelengths which is the diffraction grating efficiency $\eta_\lambda$.

The diffraction grating efficiency $\eta_\lambda$ is calibrated by measuring the intensity of the zeroth order and positive first order light. To achieve this, we place spectral bandpass filters in front of the camera at 10nm intervals, wavelengths from 440nm to 660nm. For each wavelength $\lambda$, the corresponding dispersed light projected onto a Spectralon target is captured. Figure~\ref{fig:radiometric_calibration}(d) depicts an example of the captured Spectralon target for the 500nm wavelength, showing both the positive first-order diffracted light and the zero-order light. We compute the final diffraction efficiency by deriving the intensity ratio of each positive first-order wavelength over zero-order intensity. 

\subsection{Fast Capture Synchronization}
We capture dynamic scenes by synchronizing the stereo cameras and the projector. Using this setup, we capture one group of $M$ DDSL patterns with a single black pattern, at a frame rate of 6.6 fps. Synchronization between the projector and stereo cameras is managed through the OpenGL library. The next projector pattern is preloaded in advance, and the buffer is swapped immediately after the future tasks for image capture are completed. The synchronization process is outlined in the pseudo code below.

\begin{algorithm}[H]
\caption{Pseudo-code for Multi-Camera Capture and Display}
\KwIn{Stereo camera serial numbers $serial_1$, $serial_2$.}
\KwOut{Stereo camera captured image lists.}

\SetKwBlock{Init}{Initialization}{end}
\SetKwBlock{Acquisition}{Start Image Acquisition}{end}

\Init{
    Initialize stereo cameras using serial numbers $serial_1$ and $serial_2$\;
    Configure cameras\;
    Enable software trigger mode for both cameras\;
    Initialize fullscreen display\;
    Create group of $M$ DDSL and single black pattern $Pattern\_group$\;
}

\Acquisition{
    Start image acquisition for both cameras\;
    Display the first image on the screen\;
    
    \ForEach{image $i$ in $Pattern\_group$}{
        Swap display buffers\;
        Execute software trigger for both cameras\;
        Launch threads to capture images from both cameras\;
        
        \If{$i < len(Pattern\_group) - 1$}{
            Preload the next image into the display buffer\;
        }
    }
}
\end{algorithm}

\clearpage

\section{Hyperspectral 3D Reconstruction}

\subsection{Active-stereo Depth Estimation}

We employ an active stereo imaging system to achieve accurate single-shot depth estimation using the pretrained RAFT-stereo network~\cite{lipson2021raft}. The network takes two rectified stereo images $\text{rectify}(I^i_k)$ as input and outputs a disparity map corresponding to the rectified camera view. 

\begin{figure}[h]
    \centering
    \includegraphics[width=\linewidth]{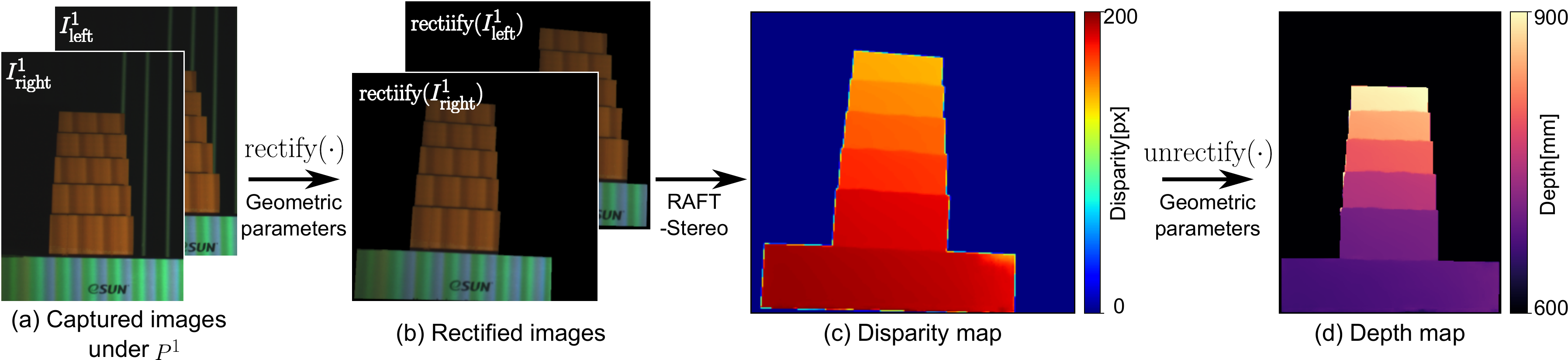}
    \caption{\textbf{Raft Stereo.} (a) Captured images under $P^1$. (b) Rectified images based on calibrated geometric paramters of stereo cameras. (c) Output of disparity map of RAFT-Stereo~\cite{lipson2021raft}. (d) Depth map.}
    \label{fig:raft_stereo}
\end{figure}

To integrate RAFT-Stereo~\cite{lipson2021raft} into our system, we first utilize the precalibrated geometric parameters to rectify the stereo image pairs before feeding them into the network. The output disparity map from RAFT-Stereo, initially aligned with the rectified view, is them transformed back to the original camera view through an inverse rectification $\text{unrectify}(\cdot)$ process. Finally, using the disparity map and the geometric calibration parameters, we compute the depth map for the target camera view. This process ensures accurate depth estimation with high fidelity, using both the state-of-the-art RAFT-Stereo framework and the precise geometric calibration of our system. Figure~\ref{fig:raft_stereo} provides an overview of this pipeline.

\subsection{Matrix-vector Multiplication}
In this section we detail the matrix-vector multiplication form as shown in Equation~\ref{eq:image_formation_matrix} of our image formation. Following describes each intensity matrix $\mathbf{I}_k$ and system matrix $\mathbf{L}$.
\begin{equation}
\label{eq:image_formation_matrix}
    \mathbf{I}_k = \mathbf{L} \mathbf{H}_k.
\end{equation}

\subsubsection{Intensity Matrix}
To construct the intensity matrix $\mathbf{I}_k$, images are captured under $M$ DDSL patterns, denoted as $\{I_k^i(p^i_k, c)\}_{i=1}^M$. To get rid of the effects of residual light and enhance reconstruction accuracy, the intensity of a black pattern image $I^B_k$ is subtracted from each captured image. This subtraction helps to remove undesired residual light intensity that may be present in the black pattern. By minimizing these residual effects, this ensures more accurate intensity measurements, contributing to improved reconstruction performance.

\subsubsection{System Matrix}
With the calibrated and refined radiometric parameters, we define the system matrix $\mathbf{L}$ as expressed as below:

\begin{align}
    \mathbf{L} =& \begin{bmatrix} \mathbf{L}_R; \mathbf{L}_G; \mathbf{L}_B \end{bmatrix}^\intercal \in \mathbb{R}^{3M \times N}, \\ 
    \mathbf{L}_c(i, j) =& \Omega^{\text{cam}}_{c, \lambda_j} \, \eta_{\lambda_j} \, \left( L^i * G \right)\left( q_{\lambda_j}^i, \lambda_j \right),
    \label{eq:system_matrix}
\end{align}
where $L_c(i,j)$ represents the system matrix element for channel $c\in\{R,G,B\}$, with $i \in \{1, \cdots, M\}$ corresponding to the i-th DDSL pattern and $j \in \{1, \cdots, N\}$ representing the target wavelength index. Here, $N$ denotes the total  number of target wavelengths. The term $L^i$ represents the light intensity of the i-th pattern, which is convolved with a Gaussian kernel $G$ with kernel size 7 and standard deviation 3 to account for optical blurring effects. This formulation ensures accurate modeling and reconstruction.

\subsection{Hyperspectral Reconstruction}
Here we outline the details of our hyperspectral optimization. Our goal is to reconstruct a highly accurate hyperspectral image $\textbf{H}_k$, using our proposed optimization method. The optimization process is performed over 1000 epochs, employing the Adam optimizer~\cite{diederik2014adam}. The learning rate is initialized at 0.05 and is reduced according to a decay schedule with a step size of 400 epochs and a decay factor of 0.5. This schedule ensures stable convergence by gradually lowering the learning rate as the optimization progresses. The proposed method is designed to refine the spectral reconstruction iteratively, achieving high accuracy across the spectral range.

\subsection{Details on Backward Model}
We describe the process for obtaining data samples and fitting our backward model as defined in Equation~\ref{eq:backward_model}. 

\begin{equation}
    \label{eq:backward_model}
    q_{\lambda} = \psi\left(\mathbf{p},\lambda\right).
\end{equation}
Here, $q_{\lambda} \in \mathbb{S}_q$ represents the projector pixel sample corresponding to a scene point $\mathbf{p} \in \mathbb{S}_{\mathbf{p}}$ and a wavelength sample $\lambda \in \mathbb{S}_{\lambda}$. The backward model $\psi$ establishes the relationship between the projector pixel coordinates and the scene geometry dependent on wavelength $\lambda$. The following subsections detail the data acquisition process used to obtain the required samples for model fitting.

\subsubsection{Data Acquisition Method}
To acquire the necessary data samples, we position a Spectralon target at five different depth positions. Using a white scanline projection, following the method proposed in~\cite{shin2024dispersed}. The scene point samples $\mathbf{p}$ are then determined using structured light triangulation method~\cite{geng2011structured}, ensuring precise 3D coordinates. 

\begin{figure}[h]
    \centering
    \includegraphics[width=\linewidth]{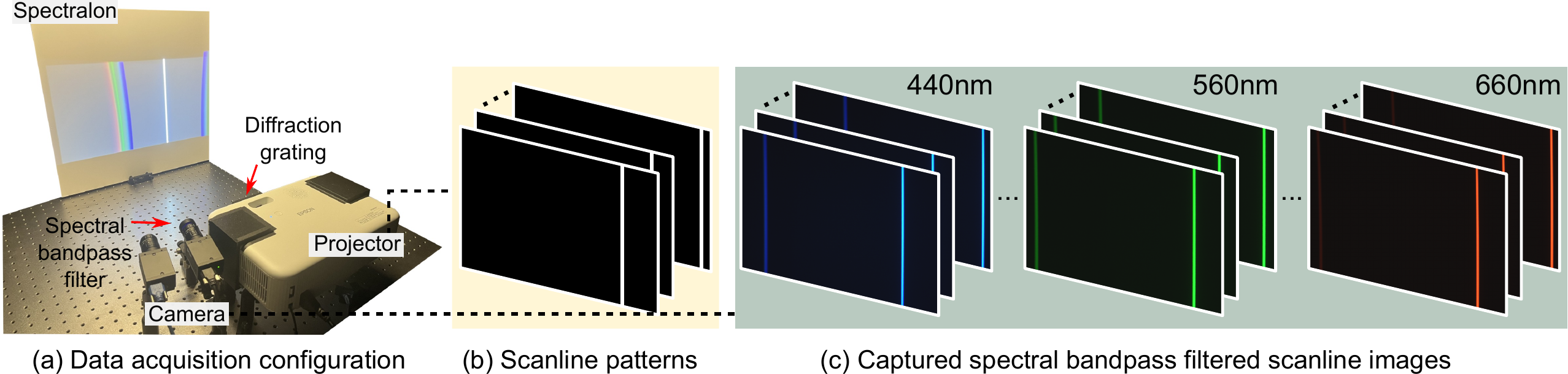}
    \caption{\textbf{Details on backward model.} (a) Data acquisition configuration system. (b) Projected white scanline patterns. (c) Captured each spectral bandpass filtered images under white scanline patterns at specific depth position.}
    \label{fig:img_formation}
\end{figure}

For each depth position, spectral data is collected by placing narrow band spectral filters in front of the camera, spanning the wavelength range of interest. Each filter captures specific wavelength $\lambda$, enabling the capture of wavelength-specific images for each scanline projection as Figure~\ref{fig:img_formation}. To determine the projector pixel $q_\lambda$ corresponding to a given wavelength $\lambda$ and scene point $\mathbf{p}$, we analyze the pixel intensity graph. These pixel intensity graph for specific wavelength $\lambda$ and pixel $\mathbf{p}$ are then modeled using a Gaussian fitting function. Previous method took the maximum value of projector coordinate whereas ours take the mean of Gaussian fitted function for sub-accurate sampling of projector coordinate $q_\lambda$. This step refines the projector coordinates $q_\lambda$ to sub-pixel accuracy, providing precise data samples for fitting the backward model.

\subsubsection{Fitting Method}
For each scene point $\mathbf{p}$ with depth coordinate $\mathbf{p}_z$, the corresponding projector pixel $q_{\lambda}$ is modeled using the following parametric equation:

\begin{equation}
\label{eq:power_function}
q_{\lambda} = \alpha \mathbf{p}_z^ {\beta} + \gamma,
\end{equation}
where $\alpha$, $\beta$, and $\gamma$ are the parameters defining the model. These parameters are estimated by fitting the collected data samples $\mathbf{p}_z$ and $q_\lambda$ to the equation using MATLAB’s non-linear power function. Additionally, linear interpolation is empolyed to estimate values for intermediate wavelength samples $\lambda$ and spatial samples $\mathbf{p}$, ensuring continuity and accuracy across the sampled range. This process forms a sub pixel accurate backward model effectively mapping each projector pixel $q_\lambda$, wavelength $\lambda$ and scene point $\mathbf{p}$.

\section{Additional Details on Black Optical Flow}
\begin{figure}[h]
    \centering
    \includegraphics[width=\linewidth]{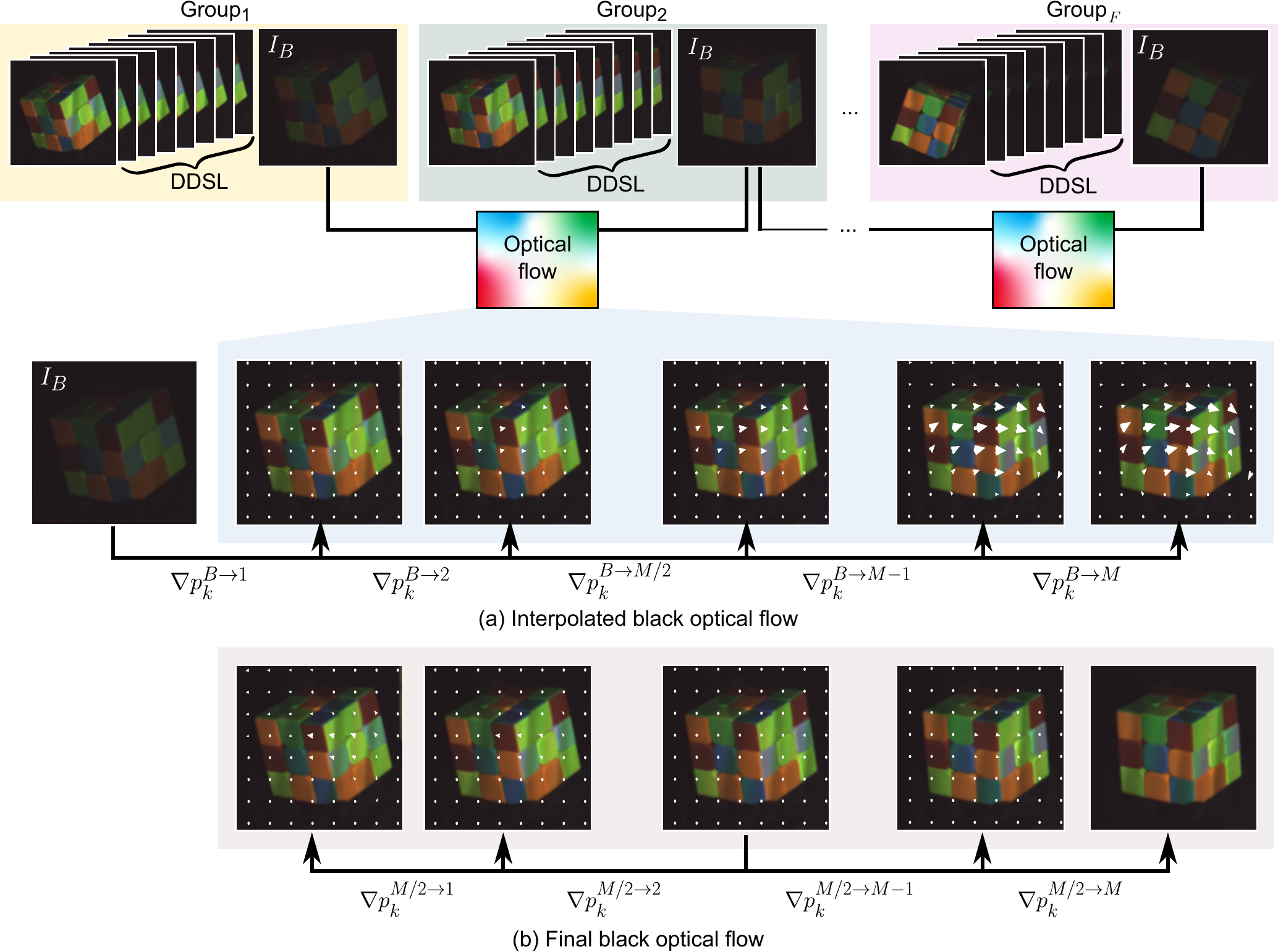}
    \caption{\textbf{Black Optical Flow.} We show the overall pipeline of black optical flow. (a) Interpolated black optical flow (b) Final black optical flow.}
    \label{fig:optical_flow}
\end{figure}

In this section we detail the black optical flow of our method. To estimate each optical flow in between black captured images $I_B$ we used pretrained network RAFT~\cite{teed2020raft}. The overall black optical flow is shown in Figure~\ref{fig:optical_flow}.

We first estimate all optical flows in between black captured images in $\{\text{Group}_i\}_{i=1}^{F}$
and all black images are multiplied with constant 3 for bright intensity cue for optical flow. As shown in Figure~\ref{fig:optical_flow}(a) with the interpolated black optical flow of a specific group, we interpolate this using cubic interpolation method and obtain the optical flow from black captured image $I_B$ to each image captured under i-th DDSL pattern $\nabla p_k^{B\to i}$. Since our goal is to obtain the target optical flow $\nabla p_k^{M/2 \to i}$ we subtract each optical flow to obtain this as the Equation below and the Figure~\ref{fig:optical_flow}(b):

\begin{equation}
    \nabla p_k^{M/2 \to i} = \nabla p_k^{B \to i} - \nabla p_k^{B \to M/2}
    \label{eq:black_optical_flow}
\end{equation}

\newpage

\section{Results on Depth Imaging}
\begin{figure}[h]
    \centering
    \includegraphics[width=\linewidth]{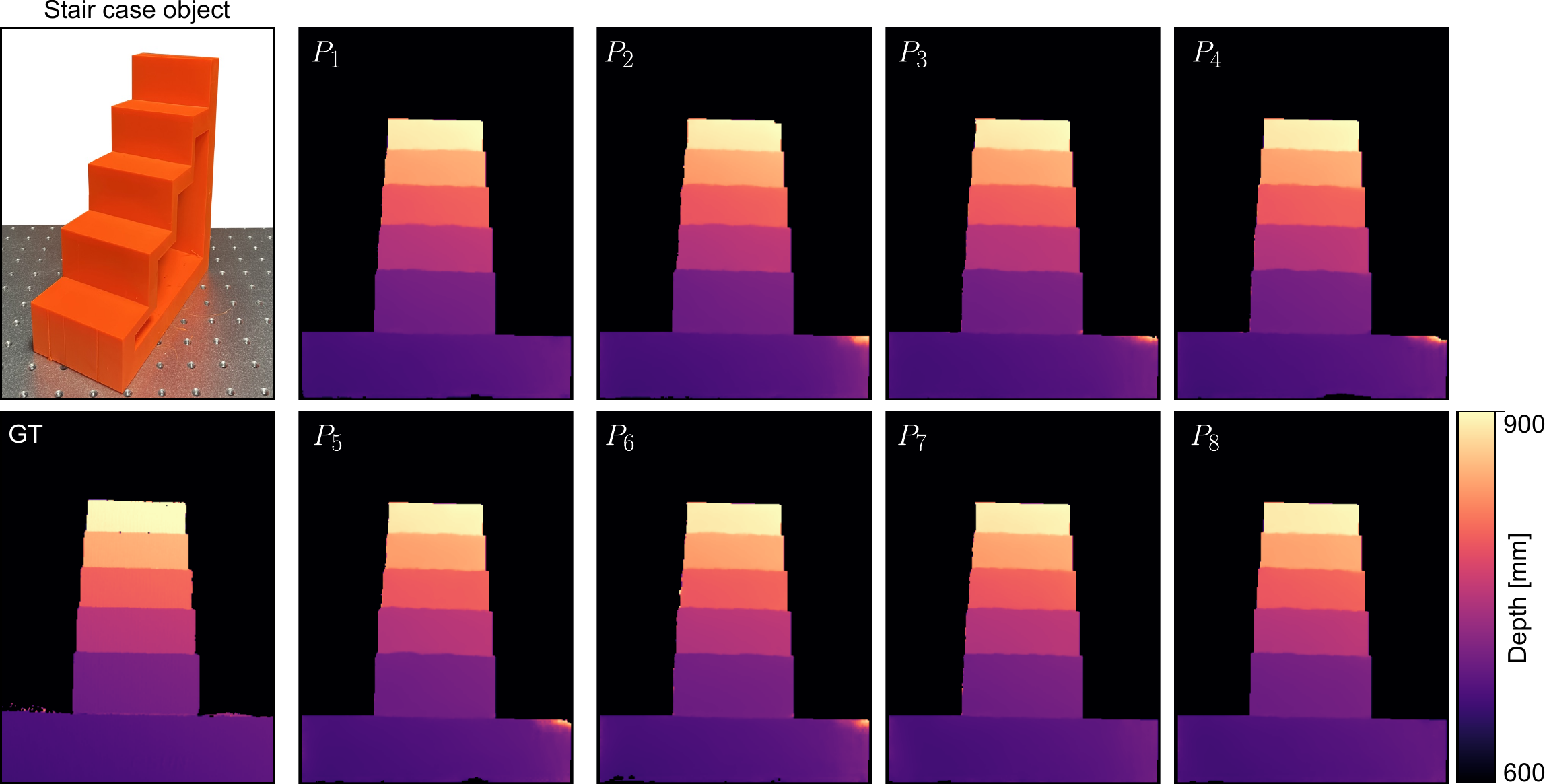}
    \caption{\textbf{Depth evaluation.} We show the depth reconstructed results for each $M$ DDSL patterns and compare it with ground truth depth earned by Structured light method~\cite{geng2011structured}.}
    \label{fig:depth}
\end{figure}

We evaluate the consistency of the reconstructed depth across $M$ DDSL patterns and compare it with the ground truth obtained using the structured light method~\cite{geng2011structured}. The results of the reconstructed depth under $M$ DDSL patterns are presented in Figure~\ref{fig:depth}, demonstrating the accuracy of our reconstruction result.

\newpage

\section{Results on Hyperspectral Imaging}
\subsection{Static Scenes}
\begin{figure}[h]
    \centering
    \includegraphics[width=\linewidth]{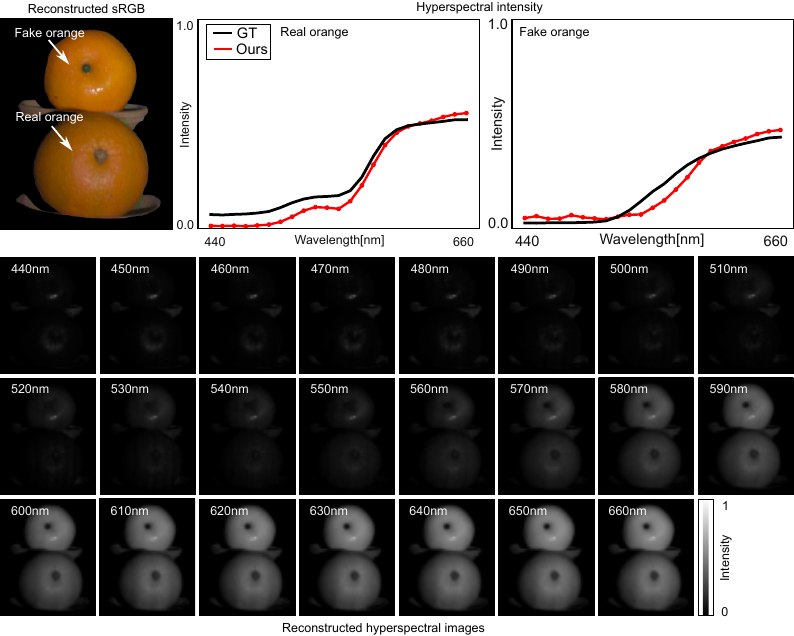}
    \caption{\textbf{Metamerism.} We show the metamerism result of a fake and real fruit(orange). Reconstructed sRGB, hyperspectral intensity with comparison with spectrometer measurements and hyperspectral images from 440nm to 660nm for 10 nm interval are depicted.}
    \label{fig:metamerism}
\end{figure}

\begin{figure}[h]
    \centering
    \includegraphics[width=0.85\linewidth]{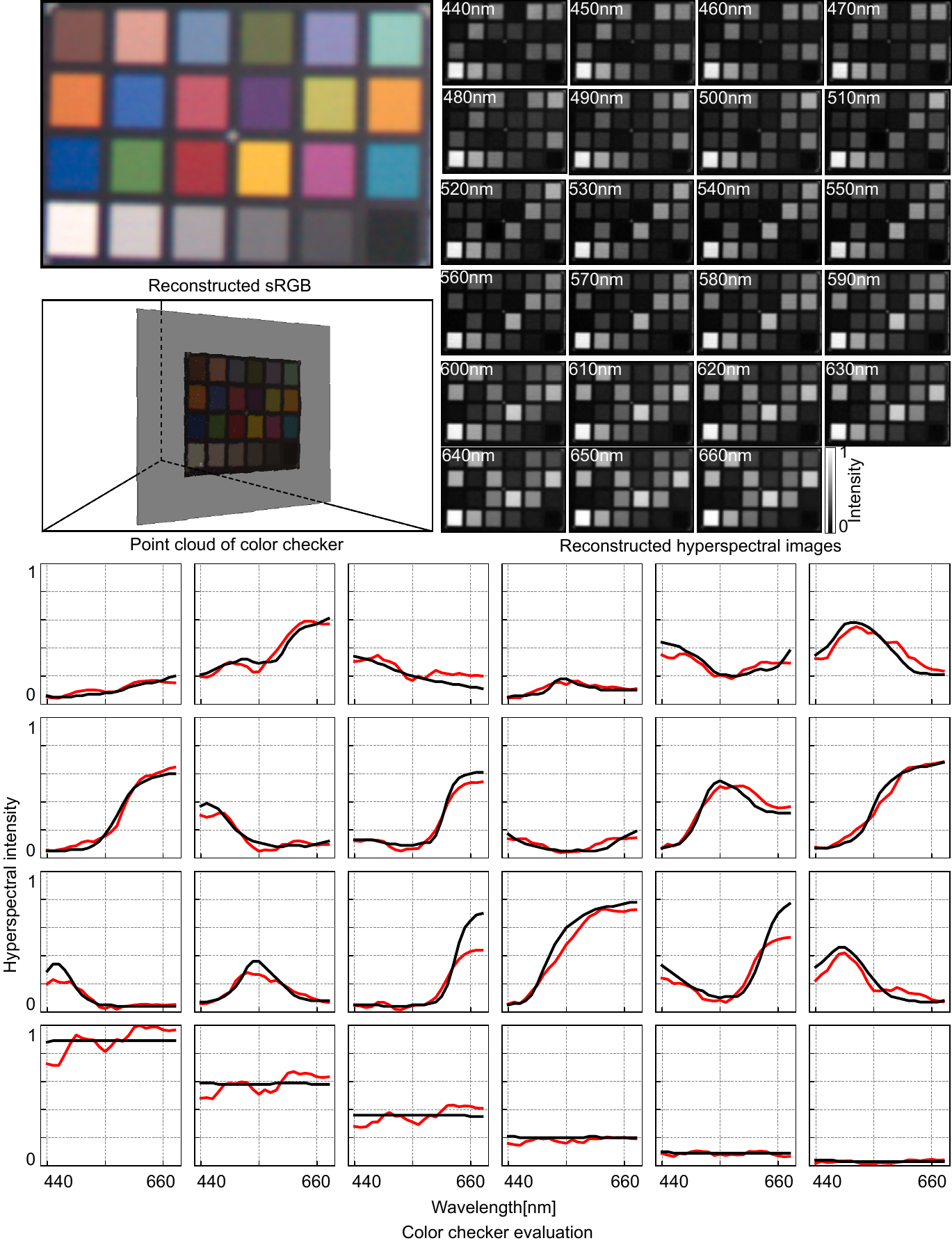}
    \caption{\textbf{Hyperspectral 3D imaging.} We show the reconstructed hyperspectral image in sRGB and reconstructed 3d point cloud, and detailed hyperspectral images from 440nm to 660nm at 10 nm interval. We show the Color Checker hyperspectral intensity comparing with ground truth spectral curves.}
    \label{fig:color_checker}
\end{figure}

\begin{figure}[h]
    \centering
    \includegraphics[width=0.7\linewidth]{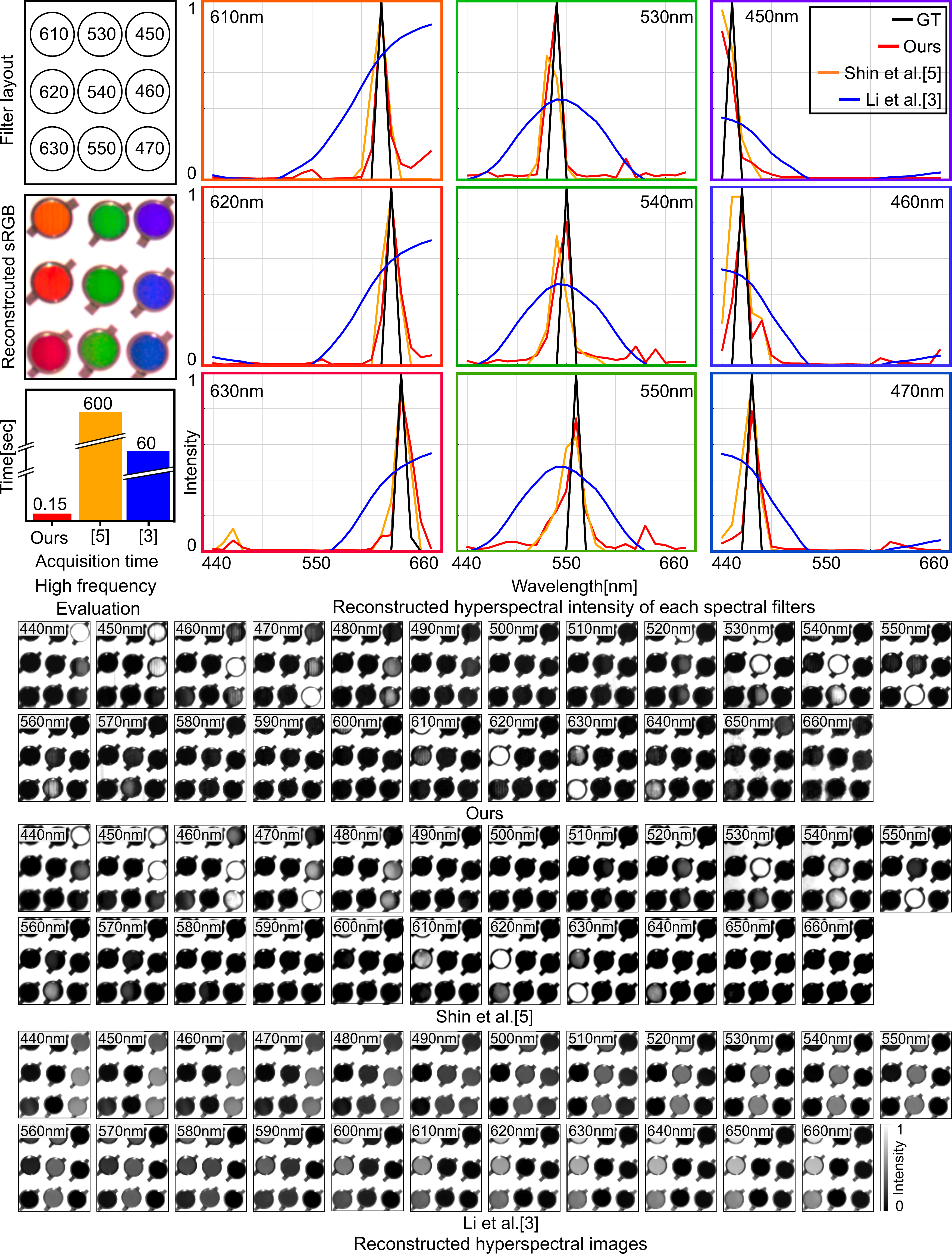}
    \caption{\textbf{High frequency evaluation.} We show the reconstructed hyperspectral images in sRGB and acquisition time of high frequency band
    pass filter scene. We compare the reconstructed hyperspectral intensity and images with Shin et al.~\cite{shin2024dispersed} and Li et al.~\cite{li2019pro}.}
    \label{fig:high_freq}
\end{figure}

We present detailed results for metamerism, ColorChecker, and high-frequency spectral curves. The reconstructed hyperspectral images, corresponding sRGB images, and spectral curves are shown for a wavelength range of 440 nm to 660 nm, sampled at 10 nm intervals. We validate the accuracy of the reconstructions with the ground truth spectral curves which were measured using a spectroradiometer. The comparison focuses on spectral curves at specific points within each scene, as illustrated in Figures~\ref{fig:metamerism}, ~\ref{fig:color_checker} and ~\ref{fig:high_freq} highlighting the accuracy of our method.

\clearpage

\paragraph{High-frequency Spectral Curves}
We compare the performance of high-frequency spectral curve reconstruction using our DDSL method with Li et al.\cite{li2019pro} and Shin et al.\cite{shin2024dispersed}. Figure~\ref{fig:high_freq} presents the reconstructed hyperspectral images produced by all three methods, covering wavelengths from 440 nm to 660 nm at 10 nm intervals. Additionally, we illustrate the acquisition time for each method, demonstrating that our DDSL method achieves a frame rate of 6.6 fps, surpassing the performance of the other approaches. Our DDSL method successfully reconstructs high-frequency hyperspectral spectral curves across all nine bandpass filters with high accuracy. In contrast, the method by Li et al.~\cite{li2019pro} struggles to resolve adjacent spectral curves, highlighting the superior reconstruction capabilities of our approach.

\clearpage

\subsection{Dynamic Scenes}
\begin{figure}[H]
    \centering
    \includegraphics[width=0.7\linewidth]{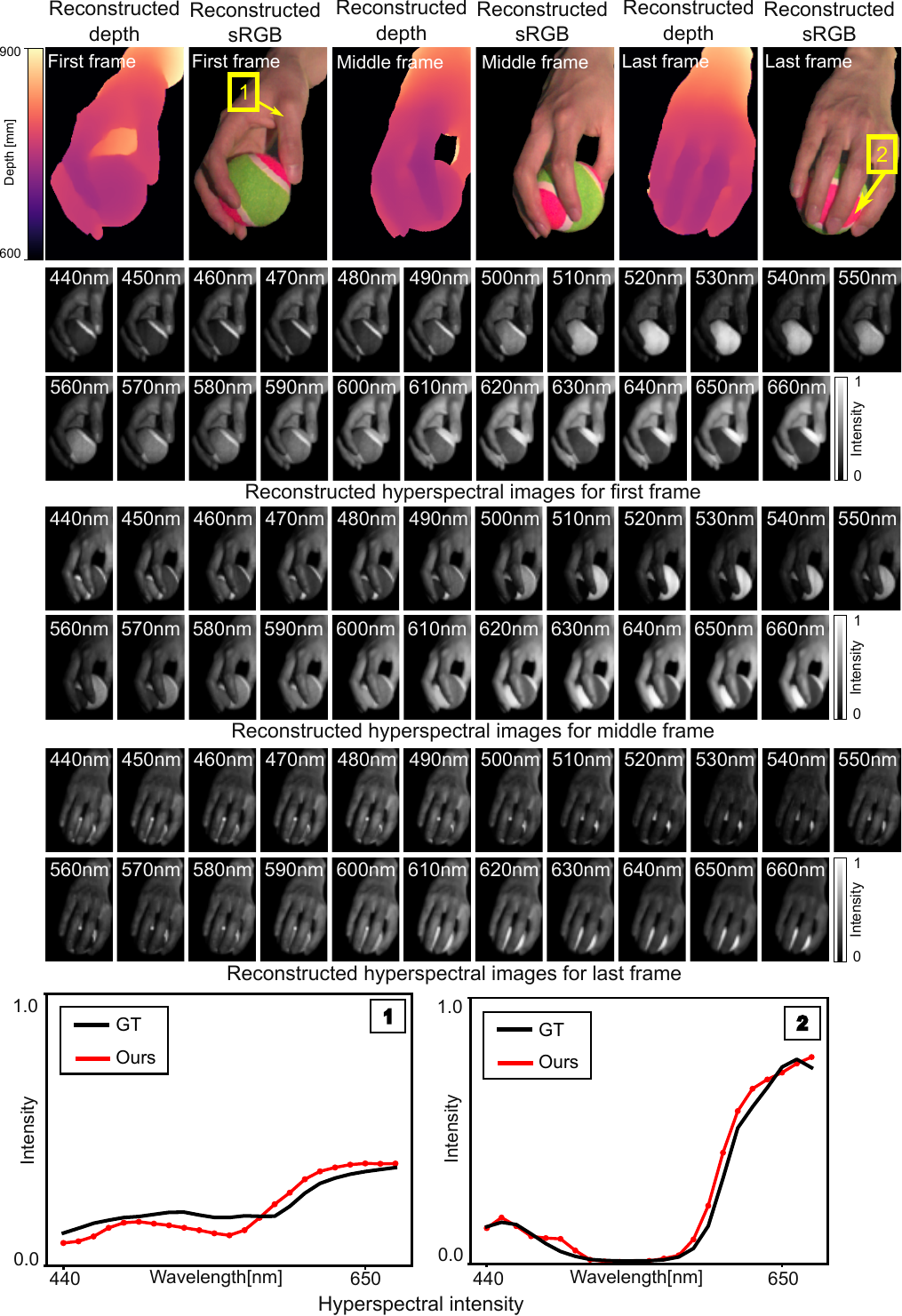}
    \vspace{-3mm}
    \caption{\textbf{Hyperspectral 3D imaging for dynamic scene.}  We show the reconstructed hyperspectral image in sRGB, reconstructed depth map, detailed reconstructed hyperspectral images from 440nm to 660nm at 10nm interval and the detailed spectral curves for specific points.}
    \vspace{-3mm}
    \label{fig:hand}
\end{figure}

\begin{figure}[H]
    \centering
    \includegraphics[width=0.8\linewidth]{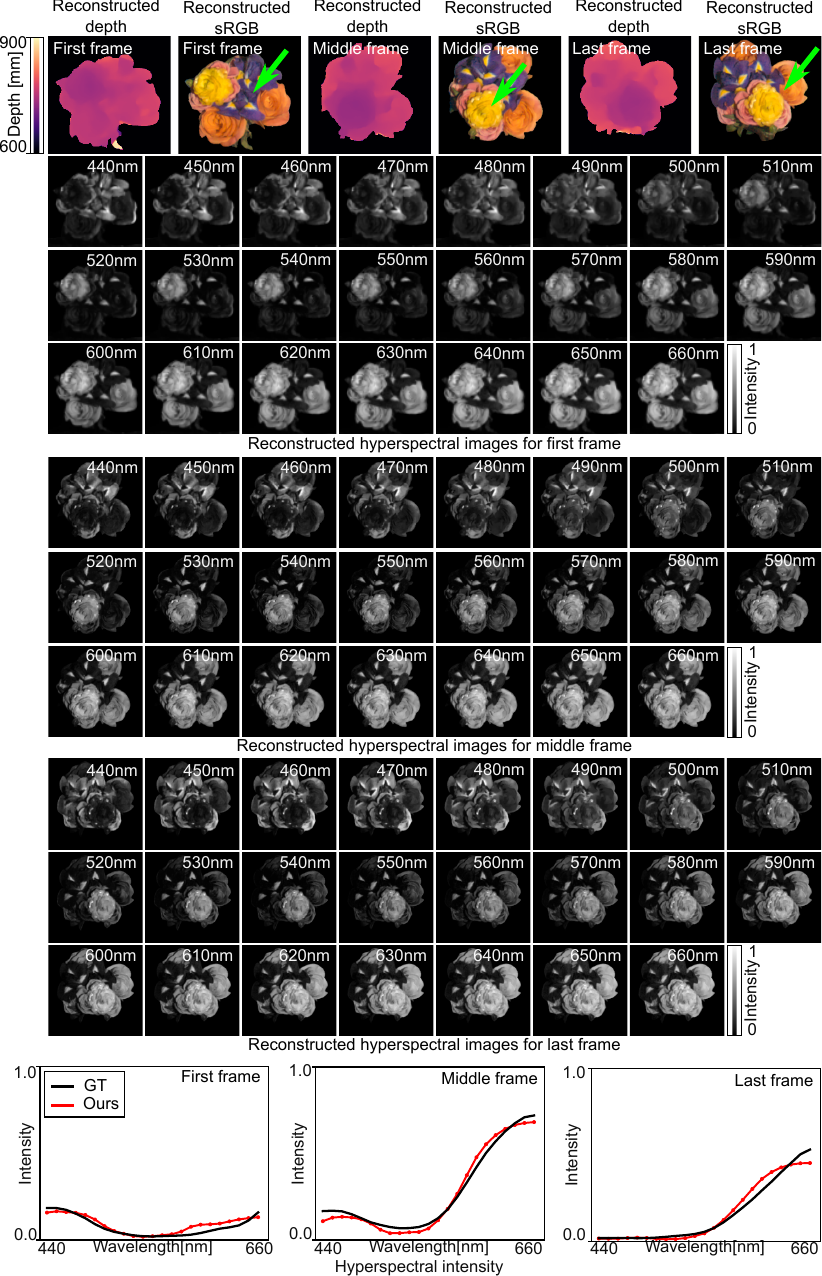}
    \caption{\textbf{Hyperspectral 3D imaging for dynamic scene.}  We show the reconstructed hyperspectral image in sRGB, reconstructed depth map, and detailed reconstructed hyperspectral images from 440nm to 660nm at 10nm interval. The detailed spectral curves are shown for specific points.}
    \label{fig:flowers}
\end{figure}

\begin{figure}[H]
    \centering
    \includegraphics[width=0.9\linewidth]{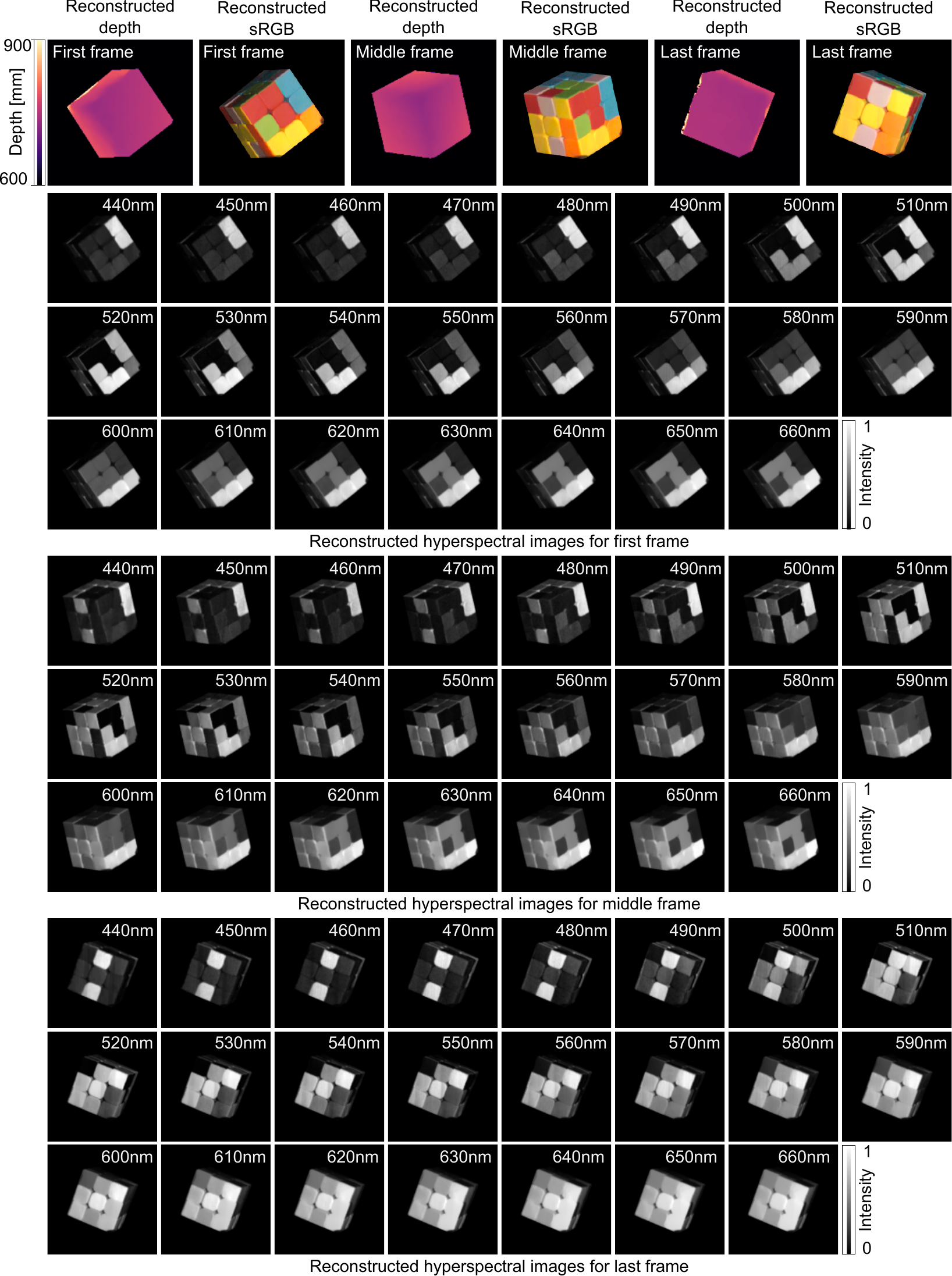}
    \caption{\textbf{Hyperspectral 3D imaging for dynamic scene.}  We show the reconstructed hyperspectral image in sRGB, reconstructed depth map, and detailed reconstructed hyperspectral images from 440nm to 660nm at 10nm interval.}
    \label{fig:cube}
\end{figure}

\begin{figure}[H]
    \centering
    \includegraphics[width=0.65\linewidth]{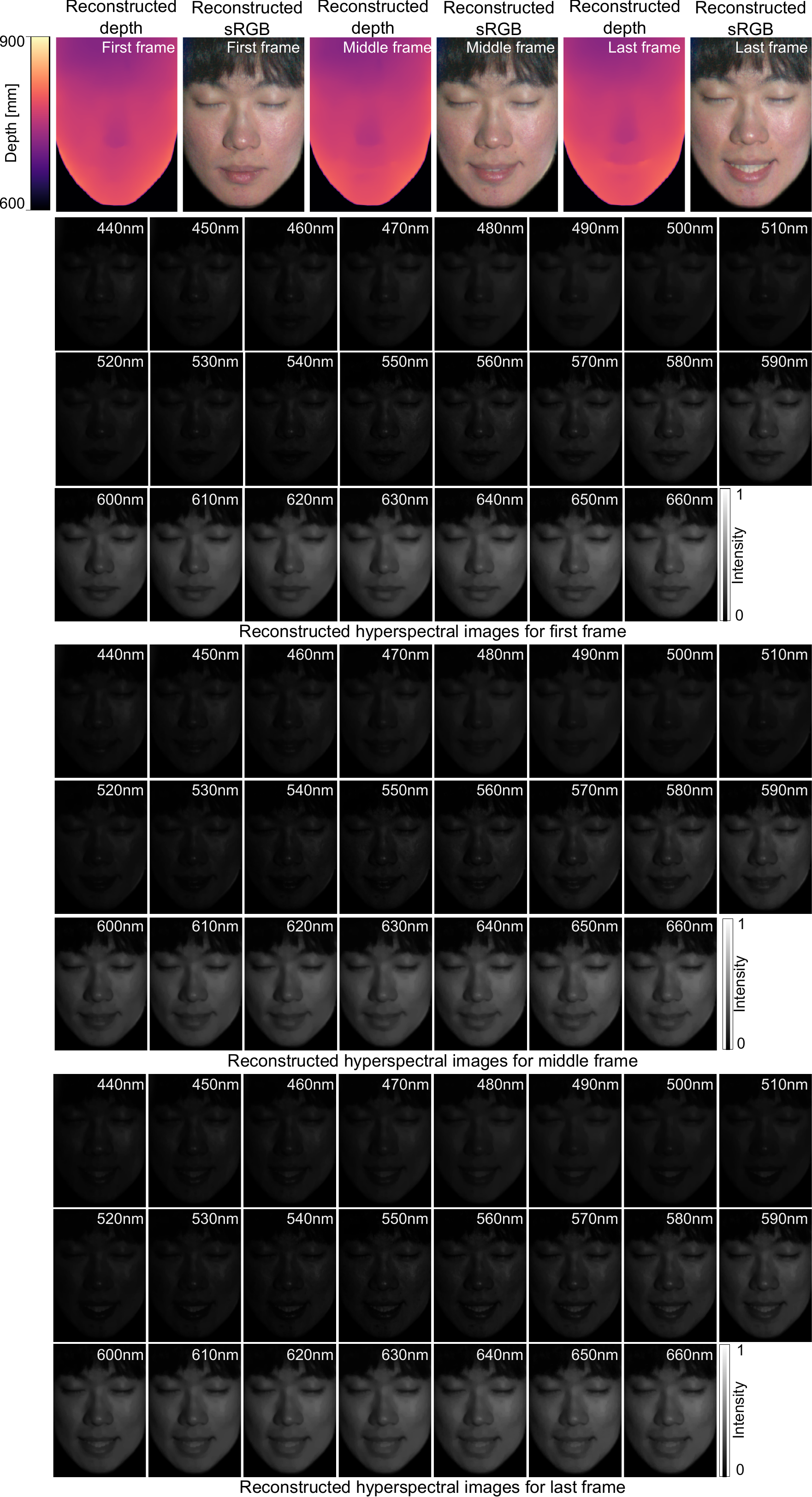}
    \caption{\textbf{Hyperspectral 3D imaging for dynamic scene.}  We show the result of face scanned reconstructed hyperspectral image in sRGB, reconstructed depth map, and detailed reconstructed hyperspectral images from 440nm to 660nm at 10nm interval.}
    \label{fig:face}
\end{figure}

We present additional results for dynamic scenes in Figures~\ref{fig:hand},~\ref{fig:flowers},~\ref{fig:cube}, and ~\ref{fig:face}. These results include reconstructed hyperspectral images in sRGB, corresponding depth maps, and detailed reconstructed hyperspectral images at 10 nm intervals. Additionally, we provide detailed spectral curves for Figures~\ref{fig:hand} and ~\ref{fig:flowers}. Face scanned result is shown in~\ref{fig:face}, depicts the hyperspectral images from 440nm to 660nm at 10nm interval. We show dynamic videos in Supplemental Video.

{\small
\bibliographystyle{ieee_fullname}
\bibliography{references_supp}
}

\end{CJK}